\documentclass[lettersize,journal]{IEEEtran}
\usepackage{amsmath,amsfonts}
\usepackage{algorithmic}
\usepackage{algorithm}
\usepackage{array}
\usepackage[caption=false,font=small,labelfont=sf,textfont=sf]{subfig}
\usepackage{textcomp}
\usepackage{stfloats}
\usepackage{url}
\usepackage{verbatim}
\usepackage{graphicx}
\usepackage{cite}
\usepackage{booktabs}  
\usepackage{multirow}  
\usepackage{hyperref}  

\begin{document}

\title{HeRB: Heterophily-Resolved Structure Balancer for Graph Neural Networks}

\author{
    \IEEEauthorblockN{Ke-Jia CHEN$^{abc*}$, Wenhui MU$^{b}$ and ZhengLiu$^{b}$\\}
    \IEEEauthorblockA{$^a$ Jiangsu Key Laboratory of Big Data Security \& Intelligent Processing, Nanjing\\ University of Posts and Telecommunications, Nanjing, 210023, Jiangsu, China\\}
    \IEEEauthorblockA{$^b$ School of Computer Science, Nanjing University of Posts and\\ Telecommunications, Nanjing, 210023, Jiangsu, China\\}
    \IEEEauthorblockA{$^c$ State Key Laboratory of Novel Software Technology, Nanjing\\ University, Nanjing, 210093, Jiangsu, China\\}
\thanks{$^*$ Corresponding author}
\thanks{$Email$ $addresses:$ \href{mailto:chenkj@njupt.edu.cn}{chenkj@njupt.edu.cn} (Ke-Jia Chen)}}

\markboth{Journal of \LaTeX\ Class Files,~Vol.~14, No.~8, August~2021}%
{Shell \MakeLowercase{\textit{et al.}}: A Sample Article Using IEEEtran.cls for IEEE Journals}

\IEEEpubid{0000--0000/00\$00.00~\copyright~2021 IEEE}

\maketitle

\begin{abstract}
Recent research has witnessed the remarkable progress of Graph Neural Networks (GNNs) in the realm of graph data representation. However, GNNs still encounter the challenge of structural imbalance. Prior solutions to this problem did not take graph heterophily into account, namely that connected nodes process distinct labels or features, thus resulting in a deficiency in effectiveness. Upon verifying the impact of heterophily on solving the structural imbalance problem, we propose to rectify the heterophily first and then transfer homophilic knowledge. To the end, we devise a method named HeRB (Heterophily-Resolved Structure Balancer) for GNNs. HeRB consists of two innovative components: 1) A heterophily-lessening augmentation module which serves to reduce inter-class edges and increase intra-class edges; 2) A homophilic knowledge transfer mechanism to convey homophilic information from head nodes to tail nodes. Experimental results demonstrate that HeRB achieves superior performance on two homophilic and six heterophilic benchmark datasets, and the ablation studies further validate the efficacy of two proposed components.
\end{abstract}

\begin{IEEEkeywords}
Graph heterophily, graph structural imbalance, knowledge transfer and graph augmentation.
\end{IEEEkeywords}

\section{Introduction}
\begin{figure}[htbp]
	\centering
	\subfloat[Node degree distribution.]
	{\includegraphics[width=0.49\linewidth]{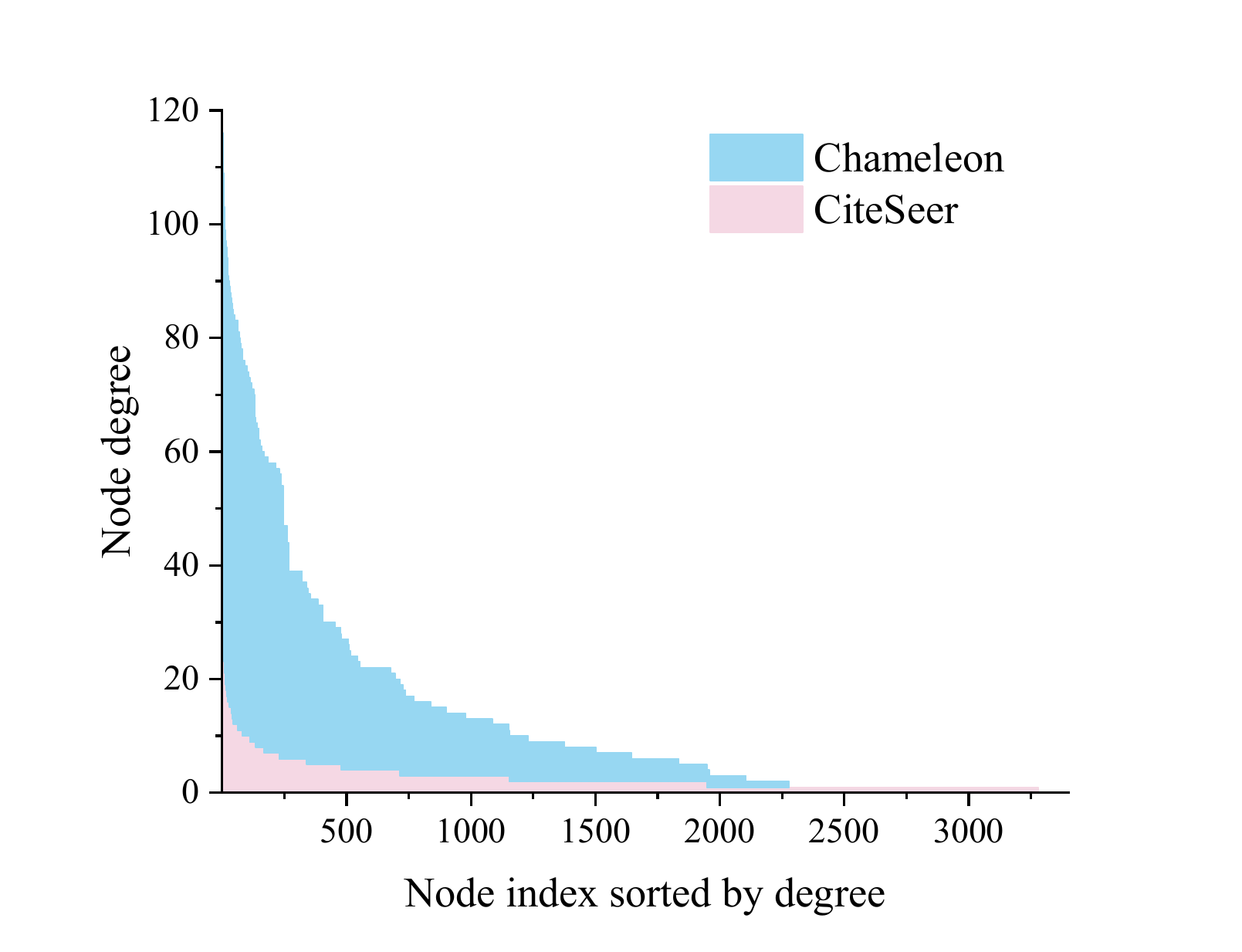}
		\label{degree}}
	\subfloat[Node classification results.]
	{\includegraphics[width=0.49\linewidth]{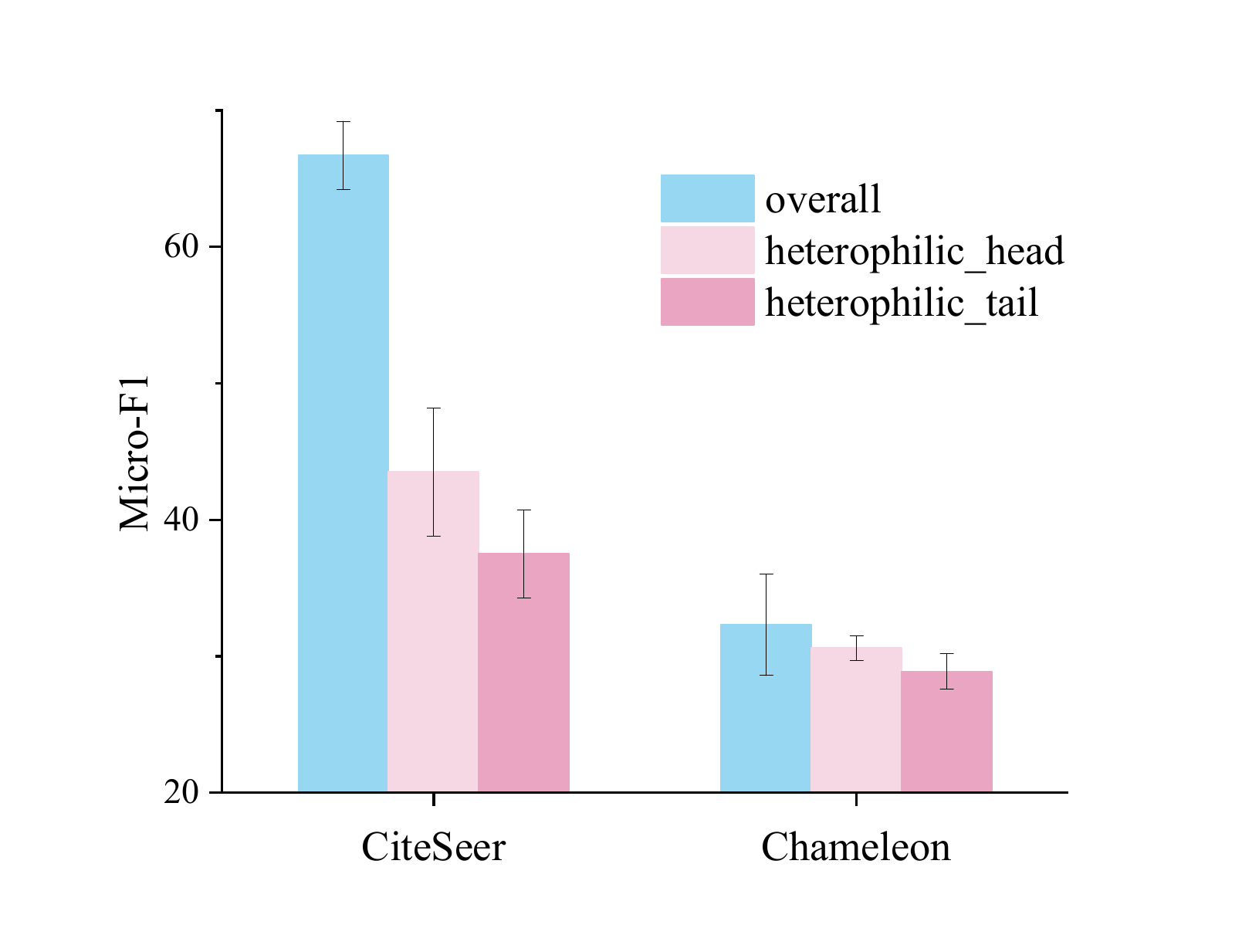}
		\label{motivation}}
	\caption{The \textbf{left} figure illustrates the imbalanced distributions of node degrees in the CiteSeer and Chameleon datasets. The \textbf{right} figure presents the Micro-F1 results of GCN on the node classification task, where both head nodes and tail nodes suffer from their heterophily, leading to poorer classification performance.}
	\label{pic}
\end{figure}

\IEEEPARstart{T}{he} popularity of Graph Neural Networks (GNNs) ~\cite{NIPS2017_5dd9db5e,Kipf2016SemiSupervisedCW,velickovic2017graph} has made the learning of more robust graph representations a prominent research topic. Despite their significant achievements, GNNs still struggle to perform well when handling graphs with extreme structural imbalance ~\cite{liu2020towards,liu2021tail}. Structural imbalance in graphs typically refers to degree imbalance, where the degrees of nodes follow a power-law distribution. As shown in Figure \ref{degree}, only a few nodes have high degrees, generally referred to as head nodes, while a significant portion of nodes have low degrees, referred to as tail nodes. Due to the imbalance of learning resources, it is difficult to achieve good generalization especially for tail nodes, making learning on graph with imbalanced structures a challenging problem.

Recent studies had made efforts to address this issue from various perspectives. Some works ~\cite{virinchi2023blade,liao2023sailor} propose to increase the number of neighbors of tail nodes through augmentation or sampling methods, some ~\cite{liu2020towards,liu2021tail} suggest transferring information from head nodes to tail nodes, while others ~\cite{feng2018representation,xia2022cengcn} aim to balance the learning process by re-weighting with degree-related weights.

\IEEEpubidadjcol

Despite their achievements, a common issue in the aforementioned works lies in the oversight of graph heterophily. Heterophily is a prevalent phenomenon in graphs, meaning that connected nodes often have different labels or features. In heterophily graphs, methods aiming to address degree imbalance may be less effective. For example, graph augmentation can alleviate degree imbalance by increasing neighbors for tail nodes but it may also exacerbate heterophily, as nearby nodes tend to have different labels. For knowledge transfer methods, transferring the neighborhood knowledge from head nodes to their heterophilic tail nodes may introduce potential noise. Additionally, some studies on heterophilic GNNs ~\cite{ma2021homophily,yan2022two} have observed that nodes with low homophily or those with high homophily but low degree fail to benefit from the message passing mechanism. Our preliminary experiments, as shown in Figure ~\ref{motivation}, also indicate that both head and tail nodes suffer from poor performance due to heterophily and tail nodes perform even worse since they have few neighobors, most of which are heterophilic.

In light of the foregoing analysis, we hold that in the study of structural imbalance, the issue of heterophily should be resolved first. Based on this observation, we propose a model named \textbf{He}terophily-\textbf{R}esolved Graph Structure \textbf{B}alancer (HeRB) which is equipped with two key components: a heterophily-lessening augmentation module and a homophilic knowledge transfer mechanism. The former mitigates heterophily through reducing inter-class edges and increasing intra-class edges, while the latter transfers homophilic neighborhood information from head nodes to tail nodes. Both components jointly achieve the objective of enriching the homophilic information of tail nodes.

The main contributions of this paper are as follows:
\begin{itemize}
  \item To the best of our knowledge, we are the first to address the structural imbalance issue in the context of graph heterophily and we propose that heterophily should be rectified first.
  \item We devise a heterophily-lessening augmentation module, which modifies the graph by minimizing heterophily and simultaneously maximizing structural proximity. Subsequently, we introduce a homophilic knowledge transfer mechanism which effectively avoids transmitting heterophilic information.
  \item Extensive experiments conducted on two homophilic and six heterophilic benchmark datasets demonstrate the effectiveness and generalizability of the proposed method.
\end{itemize}

\section{Related Works}
\subsection{Structural Imbalance-Aware GNNs}
Recent studies have recognized that structural imbalance can adversely impact the efficacy of GNNs. Some methods propose degree-aware solutions. Demo-Net ~\cite{wu2019net} and SL-DSGCN ~\cite{tang2020investigating} employ degree-specific GNNs to obtain distinct structural embeddings with varying degrees. CenGCN ~\cite{xia2022cengcn} performs edge weight adjustment and self-connections to alleviate imbalance problem for scale-free graphs. Some other methods adopt unique techniques such as knowledge transfer or distillation. Meta-tail2vec ~\cite{liu2020towards} and Tail-GNN ~\cite{liu2021tail} employ knowledge transfer mechanism. The former frames the problem as a meta-learning regression task, which has a big difference in approach compared to our method. The latter defines the scope of transfer as neighborhood knowledge transfer, but failed to take into account that the heterophilic head nodes may bring noise to tail nodes when conducting transferring, and our method just make up for this. Cold-Brew ~\cite{zheng2021cold} addresses cold-start node issue by leveraging knowledge distillation. GRACE ~\cite{xu2023grace} designs graph self-distillation and completion methods. Recently, graph augmentation is considered as one of the most effective methods to address the structural imbalance issue. SiGAug ~\cite{chen2024balancing} proposes to balance graph structure with edge augmenter and edge utility filter in signed graphs. SAug ~\cite{10.1145/3712699} designs a graph structural augmentation strategy for hub and tail nodes, respectively. SAILOR ~\cite{liao2023sailor} notices that there exists a large number of completely heterophilic tail nodes and proposes an augmentation method for these nodes.

However, existing methods solely address the issue from the perspective of graph structures, neglecting that heterophily may potentially undermine the effectiveness. In this paper, we explore the relationship between degree imbalance and heterophily. We propose that heterophilic attributes of the graph need to be rectified in advance, which is one of the primary efforts of this paper.

\subsection{Heterophily GNNs}
Various GNNs approaches ~\cite{bo2021beyond,dong2021adagnn,li2024pc} have been proposed to address the heterophily issue. Some classic works solve the problem in spatial domain using specific frameworks. H2GCN ~\cite{zhu2020beyond} proposes three effective strategies for handling heterophily. GPRGNN ~\cite{chien2020adaptive} introduces a generalized PageRank architecture to optimize the extraction of both feature and topological information. GREET ~\cite{liu2023beyond} uses an unsupervised edge-type discriminator, followed by a dual-channel graph encoder to obtain robust node embeddings. Recently, some works find that the imbalance property, along with heterophily, jointly affects model performance. In ~\cite{ma2021homophily}, when analyzing whether homophily is a necessary condition for GNNs, it is found that when the node label distribution is consistent, the larger the node degree, the better the performance of GNNs. GGCN ~\cite{yan2022two} discovers that heterophilic nodes and homophilic tail nodes perform worse than other nodes. 

However, these works mainly focus on improving the model performance in heterophilic graphs. In contrast, this paper study the impact of heterophily on solving the structural imbalance issue. Our work is not to define homophily ratio for edges or nodes but to correct the heterophilic properties of the graph, so that more targeted homophilic knowledge can be transferred.

\section{Preliminaries}
\subsection{Notations and Task Description}
$G =(V, E, A, X)$ is an undirected and unweighted graph, where $V$, $E$ are the set of nodes and edges, respectively. $A \in \{0,1\}^{N \times N}$ denotes the adjacency matrix of the graph and $N$ is the number of nodes. If there is an edge between nodes $v_i$ and $v_j$, $a_{i,j} = 1$, otherwise $a_{i,j} = 0$. $X \in \mathbb{R}^{N \times f}$ denotes the feature matrix of the graph, where $X_i \in \mathbb{R}^f$ represents the feature vector of node $v_i$ with the dimension $f$. 

Given the graph $G$, the objective of the task is to learn a mapping function: $g: A, X \rightarrow Z$, which maps each node's initial features and the graph structure to a low-dimensional space. Here, $Z \in \mathbb{R}^{N \times d}$ represents the embeddings of nodes, where $d$ is the embedding dimension.

\begin{figure*}[!t]
	\centering
	\includegraphics[width=0.95\linewidth]{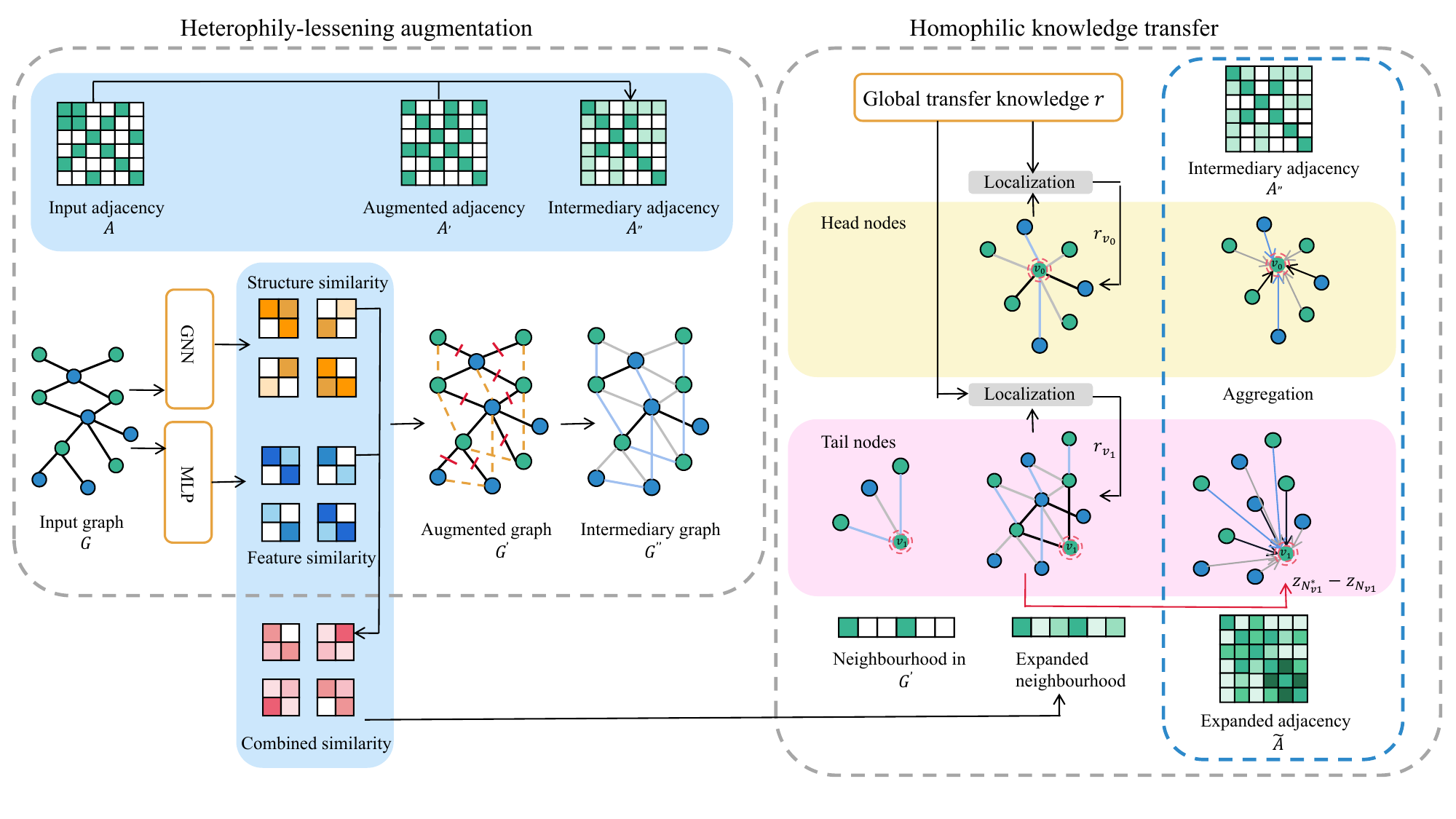}
	\caption{The overall framework of HeRB. The framework contains two modules: heterophily-lessening augmentation and homophilic knowledge transfer. The first module corrects the heterophily of the graph by increasing intra-class edges and removing inter-class edges. The second module explores the latent neighbourhood of tail nodes from head nodes and transfer the translation relationship between the expanded neighbourhood and the tail nodes during message passing.}
	\label{framework}
\end{figure*}

\subsection{Head and Tail Node Partition}
Graph structural imbalance usually refers to the power-law distribution ~\cite{achlioptas2009bias} of node degrees, where high-degree nodes are head nodes and low-degree nodes are tail nodes. To generalize to different scales of graphs, we apply the Pareto Principle ~\cite{sanders1987pareto} to categorize nodes, that is, nodes with degrees less than or equal to a threshold (we set the 80th percentile as the threshold by first sorting the node degree) are categorized as $V_{\text{tail}}$, and others are categorized as $V_{\text{head}}$. Formally, the partitioning can be expressed as:
\begin{equation}
T_{\text{head}} = \text{degree}_{80\%} = \text{Sort}(D)_{0.8}
\end{equation}
where $D$ represents the degree matrix, $\text{Sort}(D)_{0.8}$ denotes the 80th percentile value of the sorted degree sequence. Then:
\begin{equation}
V_{\text{head}} = \{ v_i \mid \deg(v_i) > T_{\text{head}} \}
\end{equation}
\begin{equation}
V_{\text{tail}} = \{ v_i \mid \deg(v_i) \leq T_{\text{head}} \}
\end{equation}
where $\deg(v_i)$ represents the degree of node $v_i$.

\subsection{Homophily Ratio}
To quantitatively measure the homophily-heterophily property of a graph, the edge-level homophily of the graph ~\cite{zhu2020beyond} is often used which can be defined as:
\begin{equation}
h(G) = \frac{\sum_{(v_i, v_j) \in E} \mathbb{I}(y_i = y_j)}{|E|}
\end{equation} 
where \(\mathbb{I}(\cdot)\) equals \(1\) when \(y_i = y_j\), and \(0\) otherwise, and \(|E|\) is the number of edges in the graph.

The edge-level homophily ratio can be further used to quantify the homophily property at the graph level. When the ratio approaches \(1\), the graph is considered strongly homophilic, and when it approaches \(0\), the graph is considered strongly heterophilic.

\section{Proposed Method}
\subsection{Heterophily-Lessening Augmentation}
Previous augmentation-based methods may be less effective due to the heterophily. To address this problem, we propose a heterophily-lessening augmentation module that uses both graph structure and node feature similarity, adding intra-class edges to capture homophilic neighborhood information and removing inter-class edges to reduce heterophilic noise.

\subsubsection{Pre-traning}
Firstly, we use a 2-layer GNN (which can be any backbone such as GCN, GAT, GraphSAGE, etc.) as an edge predictor to obtain node structural embeddings $Z_{\text{str}} \in \mathbb{R}^{|V| \times d}$, and a 1-layer MLP as a node classifier to obtain node feature embeddings $Z_{\text{fea}} \in \mathbb{R}^{|V| \times C}$. GNNs are used as the structure encoder owing to their capacity to capture node connectivity. MLP is used as the feature encoder based on the findings that MLP outperforms GNNs in node classification tasks on some heterophilic graphs ~\cite{zhu2020beyond}.

\subsubsection{Similarity Calculation}
\label{sec:similarity}
The structural embeddings $Z_{\text{str}}$ and the feature embeddings $Z_{\text{fea}}$ are first normalized to $[0, 1]$. Then, the structure similarity matrix $S_{\text{str}}$ and the feature similarity $S_{\text{fea}}$ are obtained with the following equations, respectively:

\begin{equation}
Z_{\mathbf{x}}^{ij} = \frac{Z_{\mathbf{x}}^{ij} - \min(Z_{\mathbf{x}})}{\max(Z_{\mathbf{x}}) - \min(Z_{\mathbf{x}}) + \sigma}
\end{equation}

\begin{equation}
S_{\mathbf{x}} = Z_{\mathbf{x}}^{T} Z_{\mathbf{x}}
\end{equation}
where $\sigma$ is the smoothing term ensuring that the denominator is not zero, and $\mathbf{x} \in \{str, fea\}$.

\subsubsection{Structural Perturbation}
When adding edges, for each training node, we choose $add\%$ nodes with largest $S_{\text{str}}$ value, and require that the feature similarity between nodes tends to be homophilic. The newly added edges can be represented as:
\begin{align}
E_{\text{add}} = \{ & (v_i, v_j) \in E \mid \nonumber \\
& S_{\text{str}}(v_i, v_j) \in \{\max(S_{\text{str}}(v_i, v_k)) \mid k \neq i\}, \nonumber \\
& S_{\text{fea}}(v_i, v_j) \geq T_{\text{hete}} \}. \nonumber
\end{align}
where $v_k$ represents $add\%$ nodes with the largest $S_{\text{str}}$, and $T_{\text{hete}}$ represents the heterophily threshold. This step is to exclude edges between nodes that are structurally similar but might belong to heterophilic node pairs when adding edges.

When removing edges, for each training node, we choose $remove\%$ nodes with lowest $S_{\text{str}}$ value, and require that the feature similarity between nodes tends to be heterophilic. The edges to be removed can be represented as:
\begin{align}
E_{\text{remove}} = \{ & (v_i, v_j) \in E \mid \nonumber \\
& S_{\text{str}}(v_i, v_j) \in \{ \min(S_{\text{str}}(v_i, v_m)) \mid m \neq i \}, \nonumber \\
& S_{\text{fea}}(v_i, v_j) \leq T_{\text{homo}} \}. \nonumber
\end{align}
where $v_m$ represents $remove\%$ nodes with the lowest $S_{\text{str}}$, and $T_{\text{homo}}$ represents the homophily threshold. This step is to exclude edges between nodes that are structurally far apart but might belong to homophilic node pairs when deleting edges.

The augmented adjacency matrix is denoted as $A'$. To mitigate training-testing discrepancies and overfitting, we use an intermediary adjacency matrix for subsequent knowledge transfer and message passing, which is the average of $A$ and $A'$, i.e., $A'' = \frac{A + A'}{2}$. We provide a theoretical analysis of how the augmenter addresses the heterophily problem in Appendix C.

\subsection{Homophilic Knowledge Transfer}
Given the fact that tail nodes in structural imbalanced graphs lack local homophilic neighborhood information, we propose to transfer knowledge from head nodes that exhibit higher homophily. First, we expand the neighborhood range of tail nodes to identify homophilic head nodes and establish a translation relationship, which can be approximately defined as the difference between tail node embeddings and their expanded neighborhood embeddings. During message passing, this relationship is used to bridge the gap between the expanded and observed neighborhoods, enabling knowledge transfer for tail nodes.

\subsubsection{Translation Relationship Extraction}

In homophilic graphs, there is strong similarity between a center node and its neighbors, making it easy to infer the center node's features from its neighborhood. However, in heterophilic graphs, this relational bond may not be satisfied, making inference harder. The augmentation method proposed above alleviates the heterophily, which helps to predict this relational bonds and implement reasonable neighborhood knowledge transfer, so as to improve the learning performance of tail nodes. The translation relationships between nodes and their neighborhoods in the augmented graph can be expressed as follows:
\begin{equation}
z_{v_i} + r_{v_i} \approx z_{Nv_i}
\end{equation}
where $z_{N v_i}$ represents the embeddings of all first-order neighbors of node $v_i$, after sum pooling, mean pooling, or attention pooling. $r_{v_i}$ is a learnable parameter which represents the translation relationship between node $v_i$ and its neighborhood, i.e., the knowledge to be transferred.

Head nodes have more homophilic neighborhood information after augmentation, which favours the learning of translation relationship. However, tail nodes still have limited neighborhood information. To obtain a more accurate $r_{v_i}$, more effective $z_{N v_i}$ is needed. To explore a more abundant $N_{v_i}^*$, we first expand the tail node's neighborhood range to the second-order neighbor nodes and then apply a weighted expansion. The resulting expanded neighborhood $A_{\text{expand}}$ is denoted as:
\begin{equation}
A_{\text{2-hop}} = A' \times A''
\end{equation}
\begin{equation}
A_{\text{expand}} = \alpha \times A'' + (1 - \alpha) \times A_{\text{2-hop}}
\end{equation}
where the parameter $\alpha$ is used to adjust the relative importance between the direct neighborhood and the second-order neighborhood.

In addition to expanding the tail node's neighborhood, we further prioritize selecting homophilic head nodes to learn translation relationships since head node embeddings embodies more comprehensive information. Specifically, we concatenate $Z_{\text{str}}$ and $Z_{\text{fea}}$ to obtain $Z_{\text{s\&f}}$, and compute the similarity as described in Section ~\ref{sec:similarity}. For each training tail node, the top $k$ most similar homophilic head nodes are selected within its second-order neighborhood range. A sub-adjacency matrix $A_{\text{sim}}$ is formed, with values ranging from 0 to 1. The value of 0 indicates that there is no edge between the two nodes, and the non-zero values in each row represents the hop importance between the node $v_i$ and its neighboring nodes, corresponding to the values in $A_{\text{expand}}$. In the end, the final expanded neighborhood for a tail node can be represented as follows:
\begin{equation}
\tilde{A} = \beta \times A'' + (1 - \beta) \times A_{\text{sim}}
\end{equation}
where the parameter $\beta$ adjusts the importance of homophilic head nodes during knowledge transfer operation and message passing operation. Each non-zero value in $\tilde{A}$ represents the importance score of a neighboring node in learning the translation relationship for the tail node. Close homophilic head nodes will receive higher scores, whereas distant heterophilic tail nodes will receive lower scores. For the tail nodes, the translation relationship formula can be rewritten as:
\begin{equation}
z_{v_i} + r_{v_i} \approx z_{N_{v_i}^*}
\end{equation}

\subsubsection{Translation Relationship Learning}
In order to combine global and local information during the learning of translation relationship $r_{v_i}$, we adopt a localizing strategy ~\cite{liu2021tail} which uses scaling and shifting transformations. Specifically, we use the global $r$ as background information and combine it with the node's own embedding and its neighborhood embeddings to obtain a localizing vector $r^{(l)}_{v_i}$ of the $l$-th layer, the formula can be written as:

\begin{equation}
\begin{alignedat}{2}
r_v^{(l)} &= \psi\big(z_v^{(l)}, z_{N_v}^{(l)}, r_v^{(l)}, \theta_\psi^{(l)}\big) \\
          &= \phi_\psi\big(W_\gamma^{(l),1} z_v^{(l)} + W_\gamma^{(l),2} z_{N_v}^{(l)}\big) \\
          &+ \phi_\psi\big(W_\epsilon^{(l),1} z_v^{(l)} + W_\epsilon^{(l),2} z_{N_v}^{(l)}\big)
\end{alignedat}
\end{equation}
where $\psi(\cdot)$ is the localizing function, $\theta_\psi$ is the parameters of the function, and $\phi(\cdot)$ is an activation function, $W_\ast$ are learnable parameters and $\gamma$ and $\epsilon$ represent scaling and shifting operation, respectively.

\subsection{Model Training}
\subsubsection{Message Passing}
For head nodes, their neighborhood tends to be sufficient and homophilic after the heterophily-lessening augmentation module, and the propagation can be:
\begin{equation}
\small
Z_{v_i}^{(l+1)} = g^{(l)}(A''Z_{v_i}^{(l)}) = \phi_g(A''Z_{v_i}^{(l)}W_{\theta_g}), \quad v_i \in {V_{head}}
\end{equation}
where $\phi_g(\cdot)$ is the nonlinear activation function and $W_{\theta_g}$ is the learnable parameters of the model.

For tail nodes who have insufficient neighborhood, we construct an approximately homophilic neighborhood to learn the translation relationships. To incorporate them, the embeddings of the tail nodes is updated as follows:
\begin{equation}
\small
\begin{aligned}
Z_{v_i}^{(l+1)} &= g^{(l)}(\tilde{A} Z_{v_i}^{(l)}) = g^{(l)}(\tilde{A} Z_{v_i}^{(l)} W_{\theta_g}) \\
                 &= \phi_g\big(\tilde{A}(Z_{v_i}^{(l)} + Z_{N_{v_i}^\ast}^{(l)} - Z_{N_{v_i}}^{(l)}) W_{\theta_g}\big), \quad v_i \in {V_{tail}}
\end{aligned}
\end{equation}
where $Z_{N_{v_i}}$ and $Z_{N_{v_i}^\ast}$ are the embeddings of the observed neighbors and the expanded neighbors of node $v_i$, respectively.

\subsubsection{Model Optimization}
Since the observed neighborhood of the head nodes is considered complete, the predicted difference 
$Z_{N_{v_i}^\ast}^{(l)} - Z_{N_{v_i}}^{(l)}$, which represents the discrepancy between the expected and observed neighborhoods, should be minimized. We use a loss function to enforce this tendency:

\begin{equation}
\mathcal{L}_{head} = \sum_{v_i \in \textit{head}} \sum_{l=1}^{L} \left\| Z_{N_{v_i}^\ast}^{(l-1)} - Z_{N_{v_i}}^{(l-1)} \right\|_2^2
\end{equation}

The model is trained in the semi-supervised node classification task, which can be evaluated using a cross-entropy loss function:

\begin{equation}
\mathcal{L}_{task} = \sum_{v_i} \text{CrossEnt}(Z_{v_i}^{(l)}, y_{v_i}) + \lambda \left\| \phi_g \right\|_2^2
\end{equation}
where $\lambda$ is a regularization hyperparameter, and $\phi_g$ contains all the learnable parameters of the model. The overall loss function of the model is as below:

\begin{equation}
\mathcal{L}_{overall} = \mu \mathcal{L}{head} + \mathcal{L}{task}
\end{equation}
where $\mu$ is a hyperparameter to control the importance of each constraint.

\subsection{Theoretical Analysis}
\subsubsection{Augmenter to Alleviate Heterophily Problem}
Standard GNNs for node classification consist of two steps: propagation and combination. ~\cite{li2018deeper} observed that as the number of GNN layers increases, classification performance degrades, and node embeddings eventually become dominated by node degree and initial features ~\cite{rossi2020proximity,chen2020simple}, a phenomenon known as "over-smoothing." In heterophilic graphs, connected nodes often have different labels or dissimilar features, which exacerbates over-smoothing. ~\cite{yan2022two} also pointed out a relationship between heterophily and over-smoothing, where heterophilic nodes suffer from message passing, making over-smoothing more likely. Therefore, the essence of addressing the heterophily issue is to alleviate the GNN over-smoothing problem caused by mismatched graph structures. ~\cite{shannon1948mathematical} suggests that improving the graph's topology can help alleviate GNN over-smoothing.

This work modify the graph topology through a heterophily lessening augmentation module, thereby addressing the more severe over-smoothing problem caused by graph heterophily. In the following, we will demonstrate how the augmenter increases the diversity of messages passed, reducing over-smoothing. Shannon entropy, which can describe the diversity of information, is defined as follows:
\begin{equation}
H(X) := \mathbb{E}[-\log p(X)] = - \sum_{x} p(x) \log p(x)
\end{equation}
where $x$ denotes a node's feature vector and $X$ represents the set of all possible values of $x$. 

The type of message propagated in a GNN model can be seen as the number of edges $|E|$ in the graph and the Shannon entropy of an edge in a traditional GNN can be described as:
\begin{equation}
H(G) = \sum_{i \in {E}} - p_i \log p_i
\end{equation}
where $i$ represents the index of the edge, message passed to $n_i$ nodes and total for $t_i$ times. When augmenter is used to modify the graph structure at the rate of $\delta$, the Shannon entropy of augmented graph can be described as:
\begin{equation}
E(H(\tilde{G})) = -\delta \log(\delta) + (1 - \delta) \sum_{i \in |E|} - p_i \log((1 - \delta) p_i)
\end{equation}
where $\tilde{G}$ represents the augmented graph. When $\delta \geq 0$, then $t_i \geq n_i$ and $E(H(\tilde{G})) \geq H(G)$.

With the boost of Shannon entropy, augmented graphs have more diverse information to be passed, alleviating the problem of over-smoothing. As a result, the heterophily problem can be alleviate by this.

\subsubsection{•}
Balabala.

\subsection{Time Complexity Analysis}
The proposed model has two key components: heterophily lessening augmentation module and homophilic knowledge transfer mechanism. Give a graph $G = (V,E,X,A)$, pre-training a GNN structural encoder (using GCN as an example) and a MLP feature encoder cost the time of $O(|E|d)$ and $O(Nfd)$, respectively, and $d$ is the output dimension. Computing the structure and feature similarity matrix cost the time of $O(Nf)+O(N^2f)=O(N^2f)$. The pre-training operation and similarity calculation can be conducted off-line. In the augmenter, the model sorts the nodes by structure similarity at first, costing the time of $O(N)$ when using efficient sorting strategy. For adding nodes, selecting, judgement and augmentation cost the time of $O(add\%N)$, $O(remove\%N)$ for removing nodes. The total time complexity of augmenter is $O(N(add\%+remove\%))=O(N)$. The time complexity of heterophily lessening augmentation module is $O(N)$. 

\begin{table}[h]
\centering
\caption{The statistics \lowercase{of} datasets.}
\begin{tabular}{lcccc}
\toprule
Dataset    & Nodes & Edges & Classes &$h$ \\
\midrule
Cora			& 2708	& 5429	& 7	& 0.81  \\
CiteSeer		& 3327	& 4732	& 6	& 0.74  \\
Chameleon	& 2277	& 36101	& 5	& 0.23  \\
Squirrel		& 5201	& 217073	& 5	& 0.22  \\
Actor		& 7600	& 335444	& 5	& 0.22  \\
Texas		& 183	& 309	& 5	& 0.11  \\
Cornell		& 183	& 295	& 5	& 0.30   \\
Wisconsin	& 251	& 499   & 5	& 0.21  \\
\bottomrule
\end{tabular}
\label{dataset}
\end{table}

\begin{table*}[ht]
\centering
\caption{Overall node classification performance of all methods with GCN as \lowercase{the} backbone model.}
{\footnotesize 
\begin{tabular}{lcccccccc}
\toprule
\multirow{2}*{Methods} &\multicolumn{2}{c}{Cora ($h=0.81$)} &\multicolumn{2}{c}{Citeseer ($h=0.74$)} &\multicolumn{2}{c}{Chameleon ($h=0.23$)} &\multicolumn{2}{c}{Squirrel ($h=0.22$)}  \\
\cmidrule(lr){2-3}\cmidrule(lr){4-5}\cmidrule(lr){6-7}\cmidrule(lr){8-9}
&Macro-F1 &Micro-F1 &Macro-F1 &Micro-F1 &Macro-F1 &Micro-F1 &Macro-F1 &Micro-F1\\
\midrule
MLP 		&55.2±0.8 &56.7±1.1 &51.2±1.3 &53.5±1.5 &36.0±1.2 &37.0±1.2 &27.2±0.6 &28.1±0.8\\
GCN 		&78.7±1.2 &79.2±1.2 &63.0±1.7 &66.7±2.5  &31.0±3.0 &32.3±3.7 &21.0±0.8 &21.9±0.9\\
\midrule
TailGNN\textsuperscript{*} &80.1±1.9 &80.8±2.0 &64.7±1.6 &68.3±1.6 &43.4±1.9 &43.8±1.9 &26.3±1.9 &27.9±1.0\\
GRACE &80.9±1.4 &81.6±1.4 &64.7±0.8 &70.2±0.1 &39.0±4.7 &40.1±3.4 &21.9±4.9 &26.5±3.4\\
SAug &\textbf{83.7±0.9} &\textbf{84.3±1.0} &66.4±1.6 &69.8±1.7 &40.8±1.4 &42.5±3.4 &\underline{28.8±0.8} &29.5±3.4\\
SAILOR &76.8±1.1 &78.3±0.9 &67.3±1.5 &\underline{72.9±1.6} &\underline{45.7±0.9} &46.0±1.3 &28.6±2.4 &29.4±1.0\\
\midrule
GPRGNN &\underline{81.6±0.5} &\underline{82.7±0.5} &67.1±0.7 &70.4±0.8 &25.3±3.6 &30.6±3.3 &19.7±2.9 &22.9±2.8\\
GGCN &77.9±0.7 &78.9±0.5 &61.0±1.3 &63.4±1.4 &30.7±8.3 &34.4±6.8 &20.3±4.7 &23.5±3.7\\
GREET &79.5±0.9 &80.7±0.8 &\underline{67.6±0.7} &72.4±0.5 &45.7±1.2 &\underline{46.6±1.1} &27.5±1.8 &\underline{30.6±1.7}\\
\midrule
HeRB (ours) &80.3±0.9 &81.2±0.9 &\textbf{74.0±0.8} &\textbf{75.8±0.8} &\textbf{50.6±1.5} &\textbf{51.6±1.4} &\textbf{30.7±1.5} &\textbf{32.1±1.8} \\
	&($\downarrow$3.4) &($\downarrow$3.1) &($\uparrow$6.4) &($\uparrow$2.9) &($\uparrow$4.9) &($\uparrow$5.0) &($\uparrow$1.9) &($\uparrow$1.5)\\
\midrule
\midrule
\multirow{2}*{Methods} &\multicolumn{2}{c}{Actor ($h=0.22$)} &\multicolumn{2}{c}{Texas ($h=0.11$)} &\multicolumn{2}{c}{Cornell ($h=0.30$)}
&\multicolumn{2}{c}{Wisconsin ($h=0.21$)}\\
\cmidrule(lr){2-3}\cmidrule(lr){4-5}\cmidrule(lr){6-7}\cmidrule(lr){8-9}
&Macro-F1 &Micro-F1 &Macro-F1 &Micro-F1 &Macro-F1 &Micro-F1 &Macro-F1 &Micro-F1\\
\midrule
MLP 		&25.5±0.8 &27.0±1.5 &48.1±8.4 &62.8±4.9  &49.2±12.7 &55.0±10.2 &51.1±4.5 &58.6±6.5\\
GCN         &21.1±0.9 &21.9±1.0 &29.1±6.6 &42.2±13.2 &20.5±3.1 &36.7±2.3 &19.1±2.3 &25.6±3.0\\
\midrule
TailGNN\textsuperscript{*} &20.4±5.0 &23.3±1.3 &30.2±6.0 &39.7±9.9 &43.8±16.4 &54.9±10.1 &49.8±7.2 &55.1±7.7\\
GRACE &13.4±2.3 &24.7±1.5 &35.3±8.8 &57.8±14.5 &21.9±6.9 &44.4±10.2 &26.7±4.2 &38.4±6.9\\
SAug &\underline{25.7±0.8} &26.7±1.0 &27.8±2.9 &47.2±1.9 &22.7±3.7 &32.8±5.2 &24.5±4.7 &35.2±5.1\\
SAILOR &21.2±1.8 &25.2±1.5 &32.4±6.5 &48.9±9.1 &33.8±9.3 &45.6±13.3 &29.9±4.3 &42.4±6.7\\
\midrule
GPRGNN &21.3±1.7 &25.7±1.4 &24.8±10.4 &42.5±15.6 &22.4±9.2 &53.3±5.7 &22.9±4.4 &42.0±5.8\\
GGCN &22.4±1.7 &26.3±1.2 &54.4±7.5 &71.3±5.3 &35.7±9.2 &55.3±6.6 &\underline{55.3±9.1} &\underline{60.8±7.1}\\
GREET     &\textbf{29.3±0.9} &\textbf{31.7±1.3} &\underline{55.1±4.5} &\underline{73.0±1.8} &\underline{52.8±5.5} &\underline{61.1±7.0} &46.2±3.2 &60.0±2.8\\
\midrule
HeRB (ours) &24.5±3.6 &\underline{27.3±0.5} &\textbf{55.1±4.3} &\textbf{74.1±1.4} &\textbf{77.1±2.8} &\textbf{77.3±3.0} &\textbf{58.5±7.1} &\textbf{65.7±7.4}\\
 &($\downarrow$4.8) &($\downarrow$4.4) &($+$0.0) &($\uparrow$1.1) &($\uparrow$24.3) &($\uparrow$16.2) &($\uparrow$3.2) &($\uparrow$4.9)\\
\bottomrule
\end{tabular}
}
\label{overall}
\end{table*}

Expanding neighbors to two-hop needs the time of $O(N)$. The complexity of exploring homophilic head node is $O(N)+O(Nf)+O(NlogN)=O(NlogN)$ when using locality sensitive hashing for approximate nearest $k$ neighbors. When learning the transition relationship, localizing strategy costs the time of $O(Nfd)$ and calculating of the transition relationship costs the time of $O(N\bar{d})$, where $\bar{d}$ is the average degree value of nodes in intermediary adjacency matrix. Message passing in our model costs the time of $O(|E|f)$. The time complexity of homophilic knowledge transfer mechanism is $O(NlogN)+O(Nfd)+O(N\bar{d})+O(|E|f)=O(Nfd)$ for real-world sparse datasets.

The total time complexity of HeRB is $O(N)+O(Nfd)=O(Nfd)$. This results in a total time complexity of $O(Nfd)$, reflecting linear scalability with respect to the graph size and feature dimensionality.

\section{Experiments}
\subsection{Experiment Settings}
\subsubsection{Datasets}

We conducted experiments on two public homophilc datasets (i.e., Cora and CiteSeer ~\cite{yang2016revisiting}) and six heterophlic datasets (i.e., Chameleon, Squirrel, Actor, Texas, Cornell and Wisconsin ~\cite{pei2020geom}) to compare comprehensive performance of all competitors. The statistics of these datasets can be found in Table \ref{dataset}.

The transductive semi-supervised node classification setting is adopted. As for Cora and CiteSeer, we use the public data split method provided by ~\cite{Kipf2016SemiSupervisedCW}. For other datasets, we use 10\% for validation and 20\% for test. For the training set, 20 nodes per class are used on Chameleon, Squirrel and Actor, and 5 nodes per class as well on Texas, Cornell and Wisconsin. Micro-F1 and Macro-F1  are used as the metrics.

\subsubsection{Baselines}
To validate the effectiveness of HeRB, the following methods are selected for comparation. (1) \textbf{Basic Models}: MLP ~\cite{taud2018multilayer} is the multilayer perceptron, which is a structure-agnostic model; GCN is a general spectral-based GNN that serves as the backbone for all other baselines in the experiments. (2) \textbf{Structural imbalance-aware GNNs}: SAug ~\cite{10.1145/3712699}, GRACE ~\cite{xu2023grace}, SAILOR ~\cite{liao2023sailor} and TailGNN ~\cite{liu2021tail}. We modifies the dataset splitting criteria in TailGNN, changing the task from tail node classification to overall node classification. (3) \textbf{Heterophily GNNs}: GPRGNN ~\cite{chien2020adaptive}, GGCN ~\cite{yan2022two} and GREET ~\cite{liu2023beyond}.

\subsubsection{Parameter Settings}
For all baselines, we set the layer of GNN as 2 and the hidden dimension to 32. Each method is evaluated over 10 runs, and the average performance is reported. Augmentation is applied only to the training set to avoid information leakage. We use Adam optimizer with a 0.005 weight decay and 0.7 dropout. The model is trained with a learning rate of 0.01 for 1000 epochs. Early stopping is applied with a window size of 300. For the competitors, we use the reported parameters from their papers and employ random search for unreported ones. In our method, $\lambda$ is fixed at 0.005, while other hyperparameters are tuned over using random search within the Pytorch framework. All models are implemented on two RTX4090 GPUs.

\subsection{Node Classification Performance}
The results of node classification are shown in Table \ref{overall}. The bolded data in each column is the best result and the underlined data is the second. $h$ represents the homophily ratio of the datasets, so that Cora and CiteSeer are considered as homophilic, and others are heteroohilic. The last row of each section represents the difference in performance between HeRB and the remaining optimal method. Overall, HeRB achieves optimal performance on most of the eight datasets, with an average improvement of 4.1\% in Macro-F1 and 3.0\% in Micro-F1 compared to the SOTA methods. HeRB performances particularly well on heterophilic datasets. For example, on the Cornell dataset, it achieves relative improvements of 56.6\% and 40.6\% in Macro-F1 and Micro-F1, respectively, compared to the backbone model, and 24.3\% and 16.2\% over the second-placed method.

On heterophilic datasets, the four tail-aware methods fail to achieve competitive results. It's probably because these methods directly apply knowledge transfer, distillation, or graph augmentation without considering the unique property of heterophilic graphs, potentially introducing noise. It highlights the need to correct heterophilic attributes before addressing imbalance. Moreover, HeRB consistently outperforms three heterophilic GNNs on most heterophilic datasets. It indicates that these methods may not be always effective, especially when the node degree is highly imbalanced, leaving tail nodes with insufficient learning resources. For example, GREET uses a variable-class predictor to separate high and low-frequency information, but tail nodes still lack sufficient low-frequency (homophilic) information. 

It is noticed that HeRB's performance on the Actor dataset is suboptimal, particularly in Macro-F1. We attribute this to a combination of class imbalance and sparse node features, which hinder similarity calculations and the selection of homophilic edges or head nodes. For small classes, the lack of samples and sparse features further exacerbates this issue, resulting in reduced performance.

On homophilous graph datasets, our method also outperforms the backbone model, achieving SOTA results on the CiteSeer dataset. However, the performance of HeRB on Cora is not the best. During experiments, we observed that this dataset exhibits extremely high global and local homophily, which probably invalidates our heterophily-focused design. In summary, our method demonstrates strong generalizability, effectively addressing degree imbalance issues especially on heterophilic graphs.

\subsection{Ablation Studies}

\begin{table*}[ht]
\centering
\caption{Ablation Studies on all eight datasets.}
\begin{tabular}{lcccccccc}
\toprule
\multirow{2}*{Methods} &\multicolumn{2}{c}{Cora ($h=0.81$)} &\multicolumn{2}{c}{Citeseer ($h=0.74$)} &\multicolumn{2}{c}{Chameleon ($h=0.23$)} &\multicolumn{2}{c}{Squirrel ($h=0.22$)}  \\
\cmidrule(lr){2-3}\cmidrule(lr){4-5}\cmidrule(lr){6-7}\cmidrule(lr){8-9}
&Macro-F1 &Micro-F1 &Macro-F1 &Micro-F1 &Macro-F1 &Micro-F1 &Macro-F1 &Micro-F1\\
\midrule
GCN 		&78.7±1.2 &79.2±1.2 &63.0±1.7 &66.7±2.5  &31.0±3.0 &32.3±3.7 &21.0±0.8 &21.9±0.9\\
\midrule
$A$ &78.6±1.0 &79.4±1.5 &62.6±1.5 &66.8±2.4 &43.1±3.0 &40.8±1.3 &26.0±1.5 &27.2±1.6\\
$A_{he}$ &78.8±0.7 &79.4±1.0 &63.2±0.9 &67.2±1.2 &46.5±1.7 &46.5±1.4 &27.2±1.1 &27.8±0.8\\
$B$ &79.2±0.8 &79.7±0.9 &69.8±1.3 &72.3±2.1 &31.8±0.9 &35.3±1.9 &23.6±0.7 &27.8±0.5\\
$B_{ho}$ &79.8±0.7 &80.6±1.0 &70.6±1.5 &72.9±1.4 &44.4±1.2 &44.8±1.5 &28.6±1.1 &29.3±0.5\\
\midrule
$A_{he}B_{ho}$ (HeRB) &80.3±0.9 &81.2±0.9 &74.0±0.8 &75.8±0.8 &50.6±1.5 &51.6±1.4 &30.7±1.5 &32.1±1.8 \\
\midrule
\midrule
\multirow{2}*{Methods} &\multicolumn{2}{c}{Actor ($h=0.22$)} &\multicolumn{2}{c}{Texas ($h=0.11$)} &\multicolumn{2}{c}{Cornell ($h=0.30$)}
&\multicolumn{2}{c}{Wisconsin ($h=0.21$)}\\
\cmidrule(lr){2-3}\cmidrule(lr){4-5}\cmidrule(lr){6-7}\cmidrule(lr){8-9}
&Macro-F1 &Micro-F1 &Macro-F1 &Micro-F1 &Macro-F1 &Micro-F1 &Macro-F1 &Micro-F1\\
\midrule
GCN         &21.1±0.9 &21.9±1.0 &29.1±6.6 &42.2±13.2 &20.5±3.1 &36.7±2.3 &19.1±2.3 &25.6±3.0\\
\midrule
$A$ &22.1±1.1 &23.4±2.0 &23.0±4.1 &48.3±5.7 &28.6±3.8 &40.0±5.6 &39.2±4.9 &55.8±2.9\\
$A_{he}$ &22.3±1.1 &24.0±2.3 &23.6±5.1 &50.8±1.9 &30.6±2.4 &43.3±4.0 &45.8±2.5 &58.2±3.3\\
$B$ &23.0±1.0 &24.5±0.8 &28.0±8.3 &37.3±5.0 &49.3±14.2 &57.5±9.4 &34.0±13.1 &40.9±15.6\\
$B_{ho}$ &22.5±0.7 &24.2±1.3 &31.8±7.6 &43.1±9.8 &52.1±11.9 &59.1±7.4 &47.2±27.6 &56.5±26.6\\
\midrule
$A_{he}B_{ho}$ (HeRB) &24.5±3.6 &27.3±0.5 &55.1±4.3 &74.1±1.4 &77.1±2.8 &77.3±3.0 &58.5±7.1 &65.7±7.4\\
\bottomrule
\end{tabular}
\label{ablation}
\end{table*}

To observe the effectiveness of each module in HeRB, the ablation study is conducted on all eight datasets. Table \ref{ablation} shows the results, where $A$ represents general augmentation without reducing heterophily and $A_{he}$ represents heterophily-lessening augmentation; $B$ represents traditional knowledge transfer and $B_{ho}$ represents homophilic knowledge transfer; $A_{he}B_{ho}$ represents both modules are used, i.e., the complete model.

Traditional augmentation method, which only using structural embedding to compute similarity, has inferior performance to heterophily lessening augmentation module in the all datasets. On Cora and CiteSeer, traditional augmentation method even decrease the performance of backbone model. Since these datasets inherently exhibit high homophily, where any heterophilic noise can degrade the model's performance. In contrast, the heterophily lessening augmentation module consistently improves performance, as it takes into account the homophilic and heterophilic properties of edges and augments accordingly.

The conventional head-tail knowledge transfer strategy does improve model performance, but on some heterophilic graph datasets such as Texas, directly transferring neighbors' information introduces heterophilic noise, which harms the model’s performance. Our proposed homophilic knowledge transfer mechanism, which assigns homophilic head nodes and emphasizes the transfer of higher weights, effectively addresses this issue and results in improvements across all datasets.

Overall, each component of our model plays a significant role. When the two modules are combined, they can achieve the objective of addressing the degree imbalance problem of heterophilic graphs.

\subsection{Parameter Sensitivity Analysis}
In this section, we present the sensitivity analysis of the key parameters in the model. Specifically, the parameter $T_{hete}$ represents the non-heterophilic threshold for feature similarity when adding edges, while $T_{homo}$ denotes the non-homophilic threshold for removing edges. The parameter $\alpha$ controls the significance of information from different hops of neighbors, whereas $\beta$ determines the contribution weight of homophilic heads during knowledge transfer. Additionally, $k$ specifies the range of exploration for homophilic head neighbors, and $\mu$ reflects the importance of constraints on head nodes.

\begin{figure*}[htbp]
	\centering
	\subfloat[Cora]
	{\includegraphics[width=0.24\linewidth]{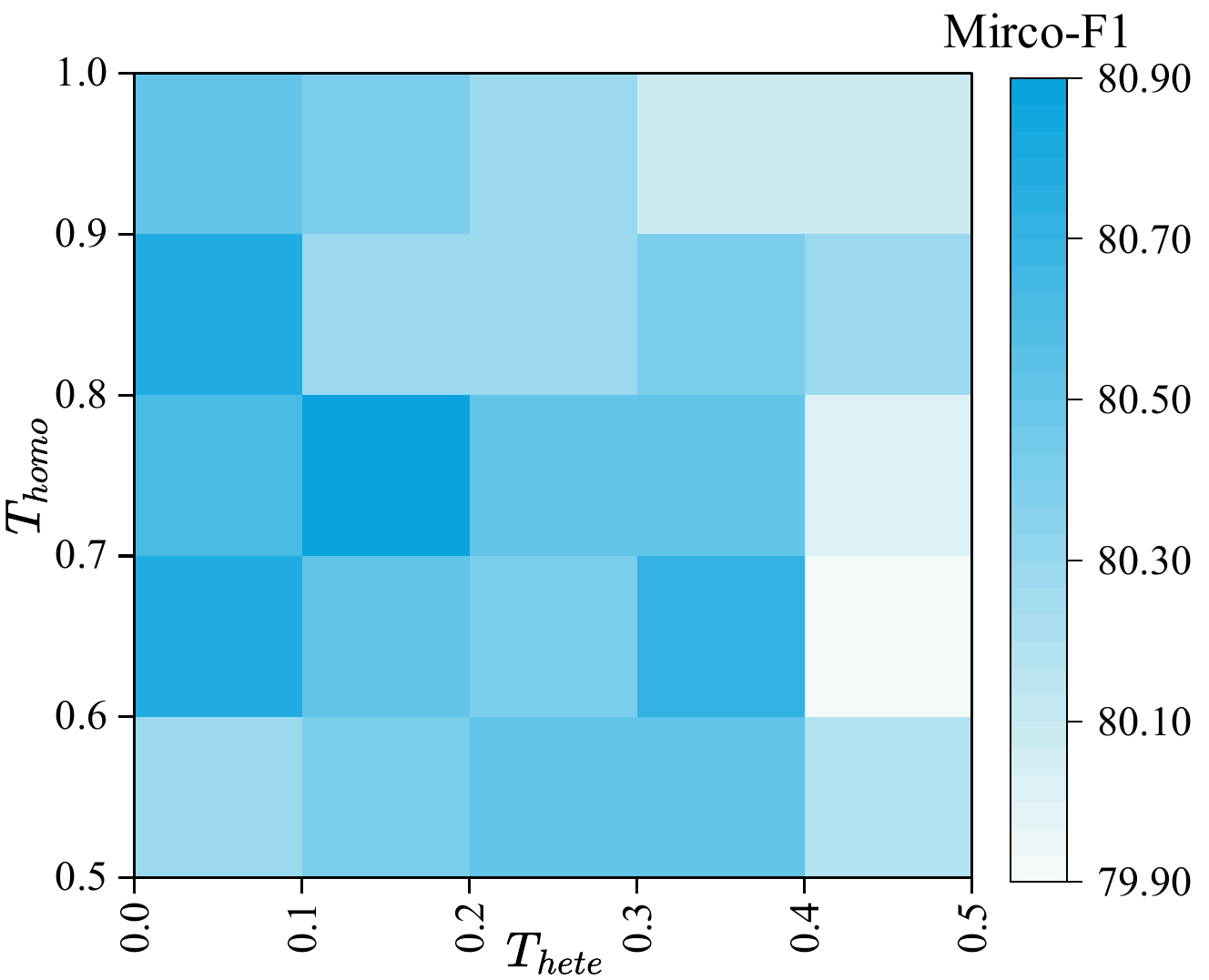}
		\label{Cora_thres}}
	\subfloat[CiteSeer]
	{\includegraphics[width=0.24\linewidth]{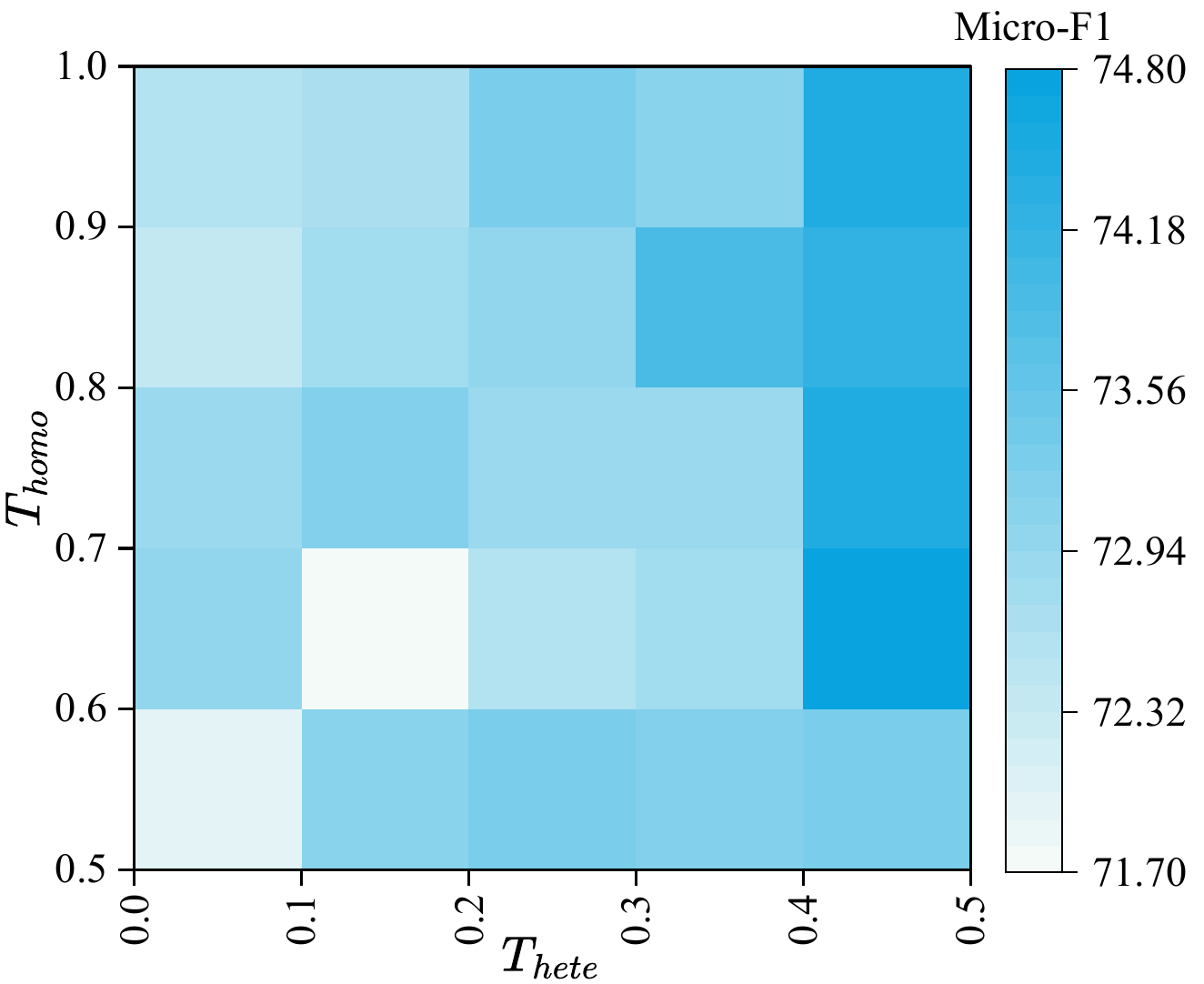}
		\label{citeseer_thres}}
	\subfloat[Chameleon]
	{\includegraphics[width=0.24\linewidth]{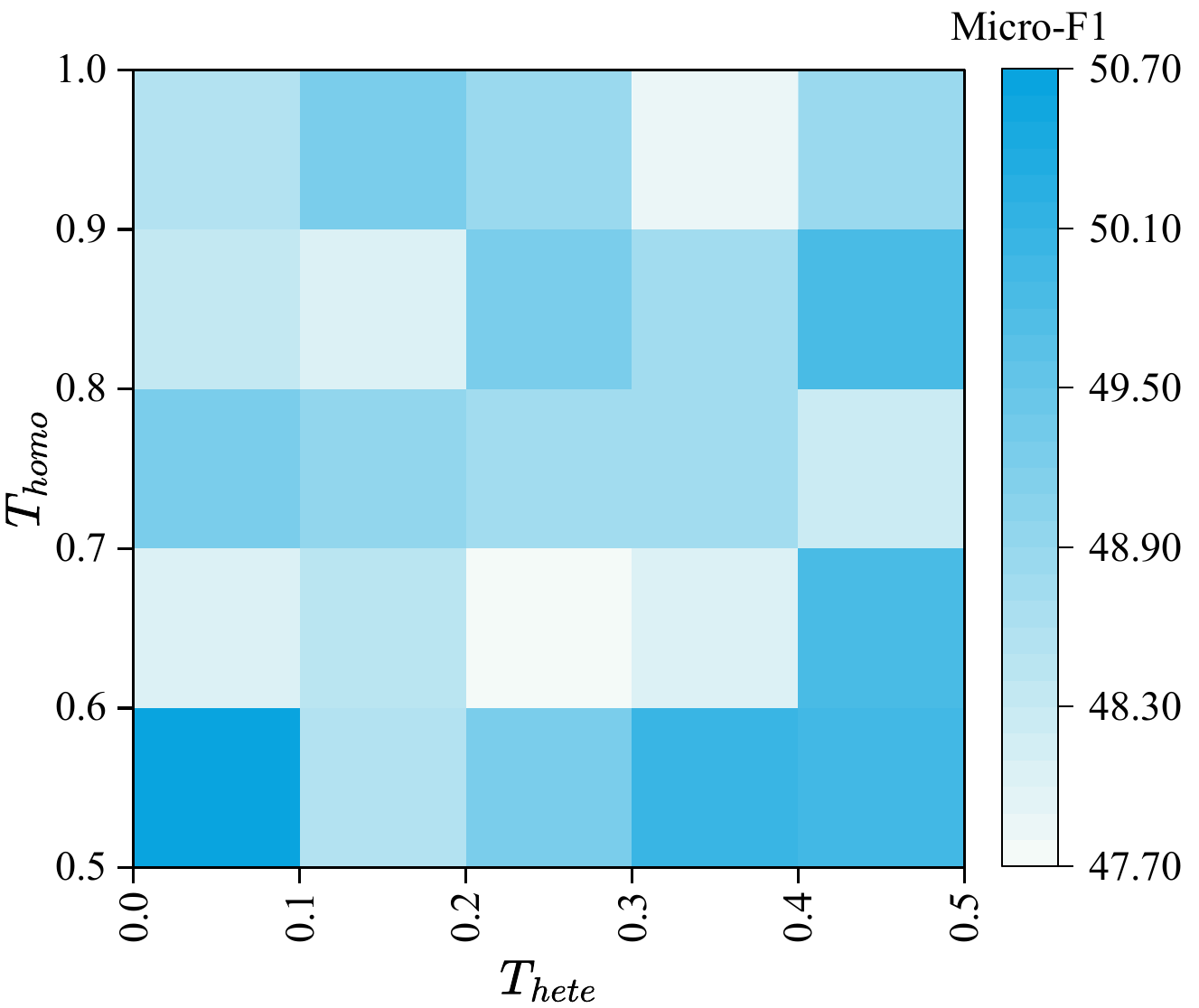}
		\label{chameleon_thres}}
	\subfloat[Squirrel]
	{\includegraphics[width=0.24\linewidth]{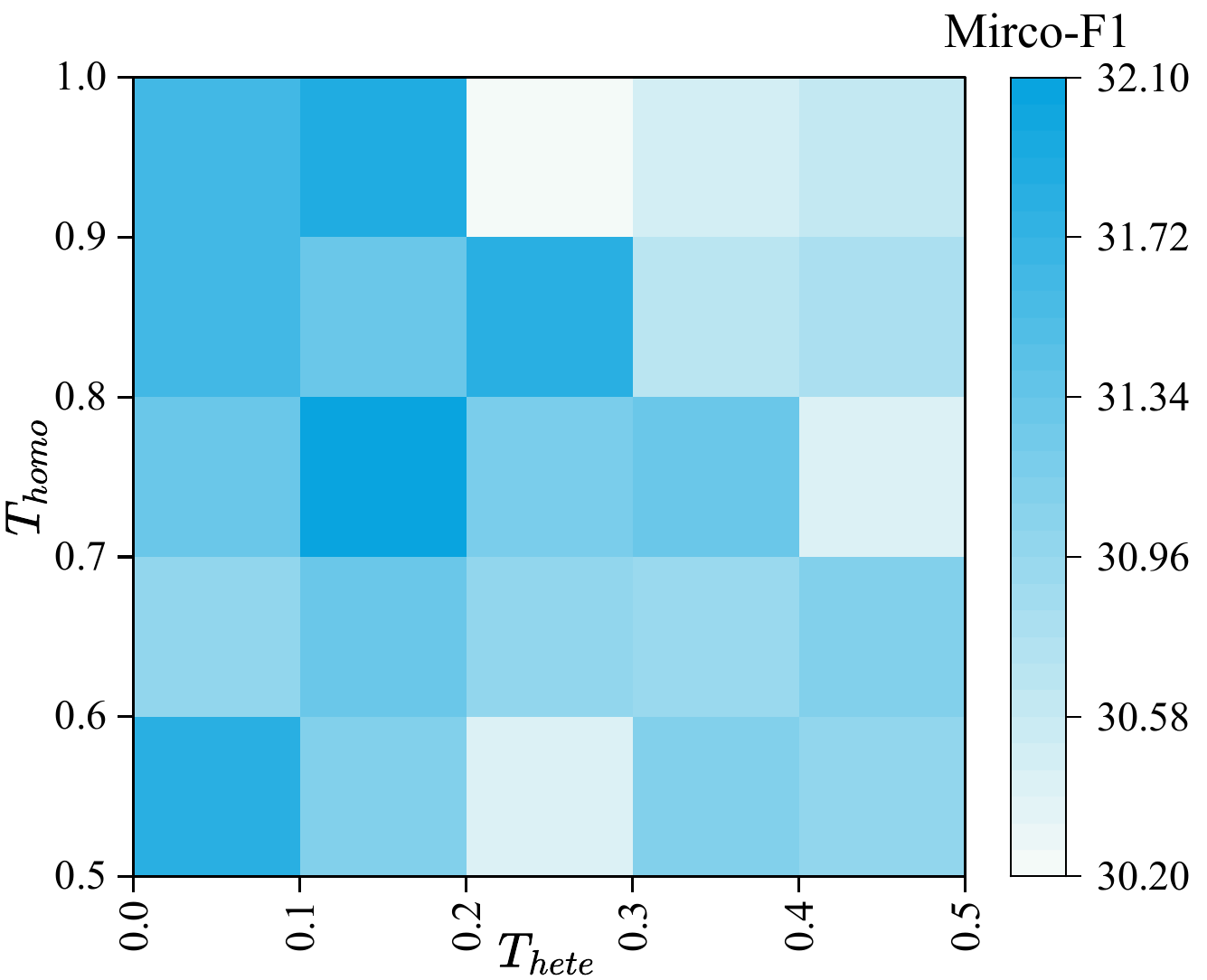}
		\label{squirrel_thres}}
		
	\subfloat[Actor]
	{\includegraphics[width=0.24\linewidth]{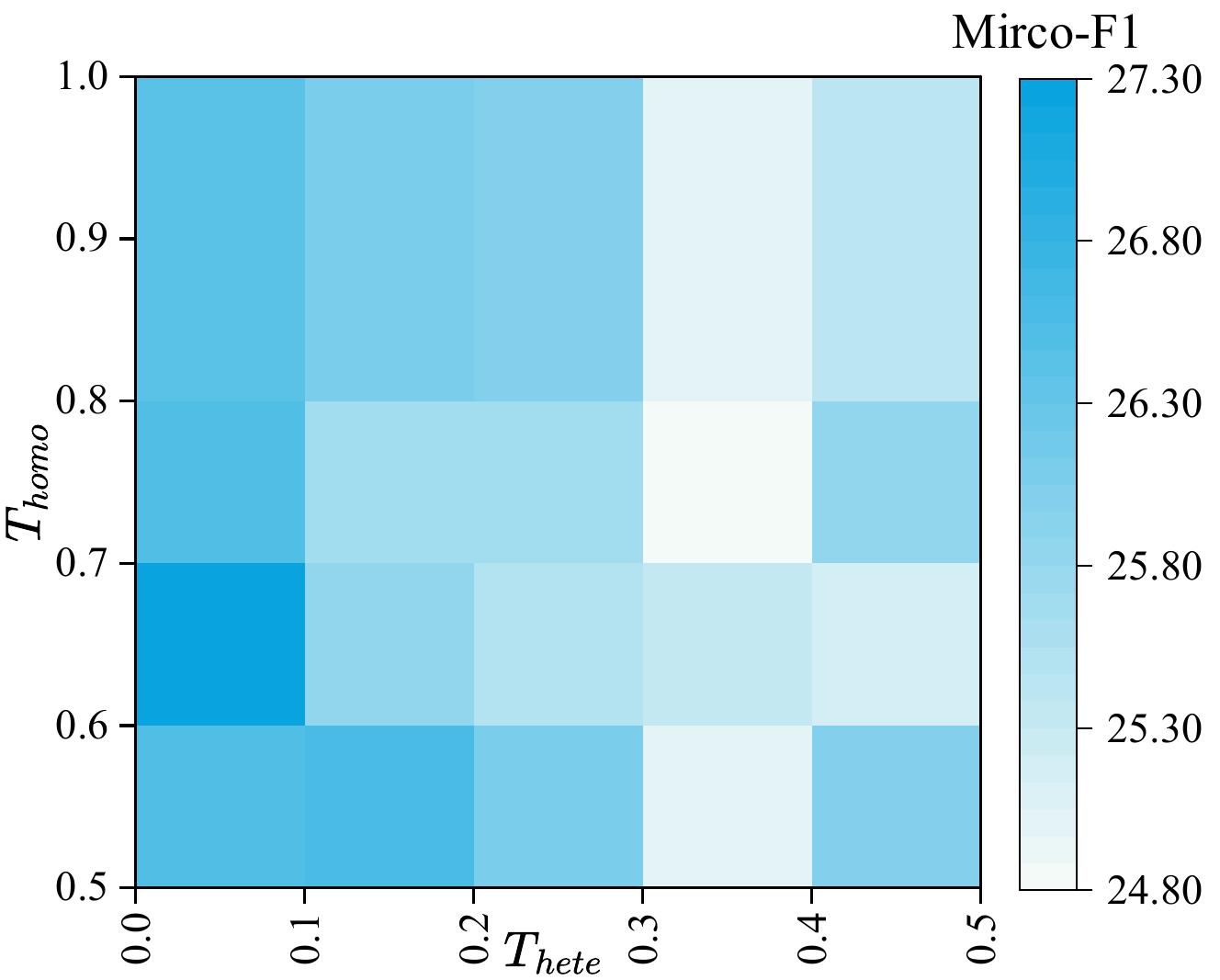}
		\label{actor_thres}}
	\subfloat[Texas]
	{\includegraphics[width=0.24\linewidth]{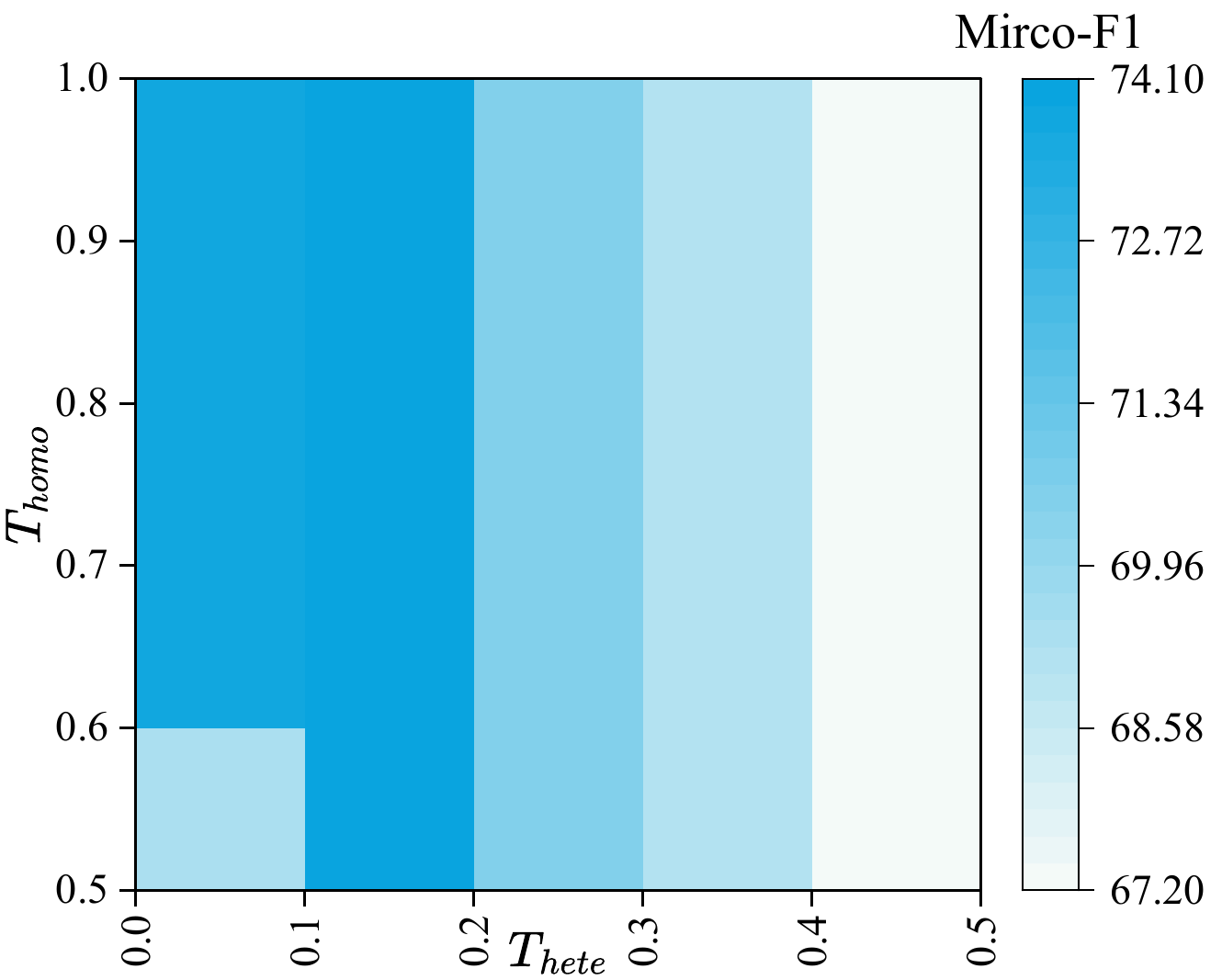}
		\label{texas_thres}}
	\subfloat[Cornell]
	{\includegraphics[width=0.24\linewidth]{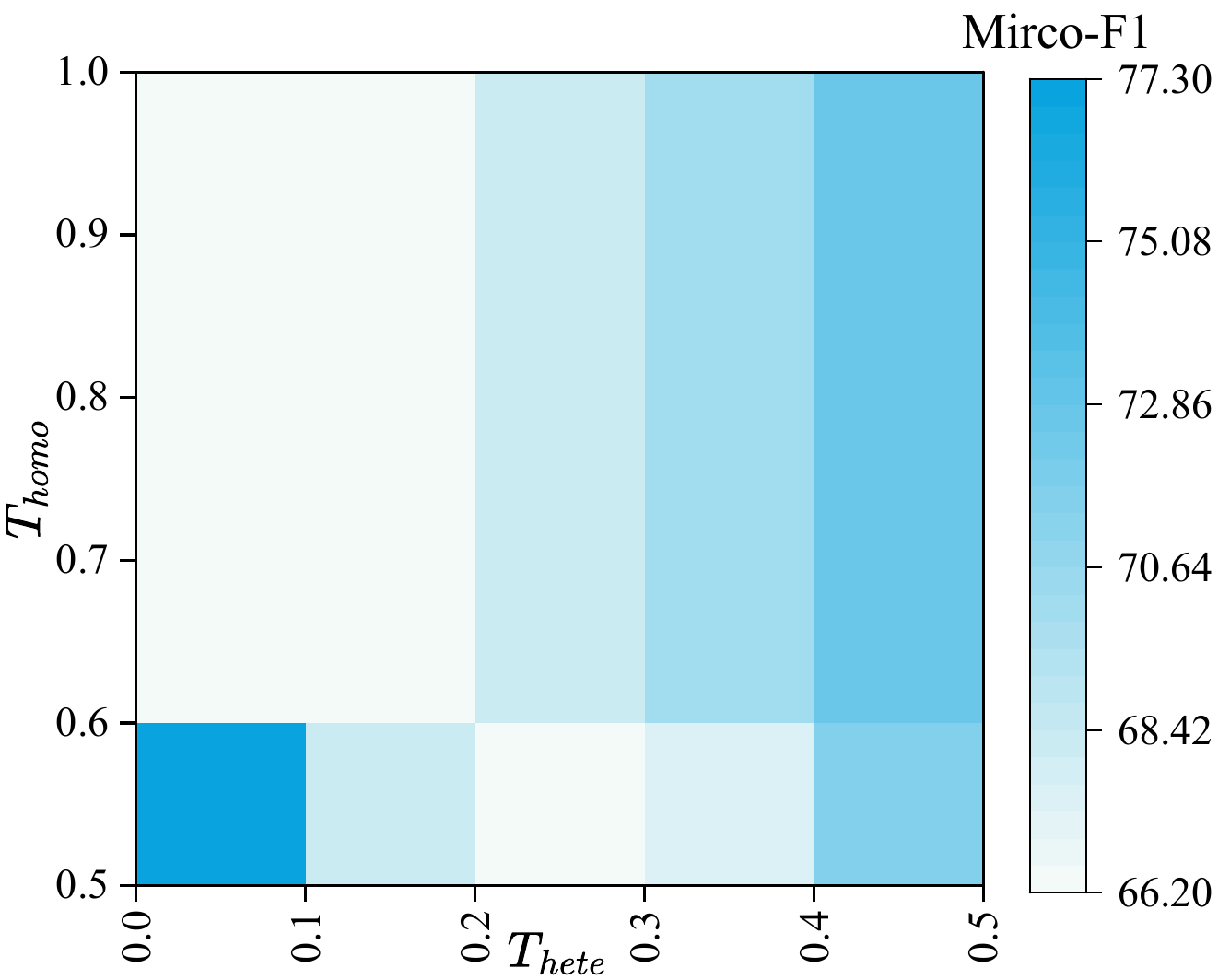}
		\label{cornell_thres}}
	\subfloat[Wisconsin]
	{\includegraphics[width=0.24\linewidth]{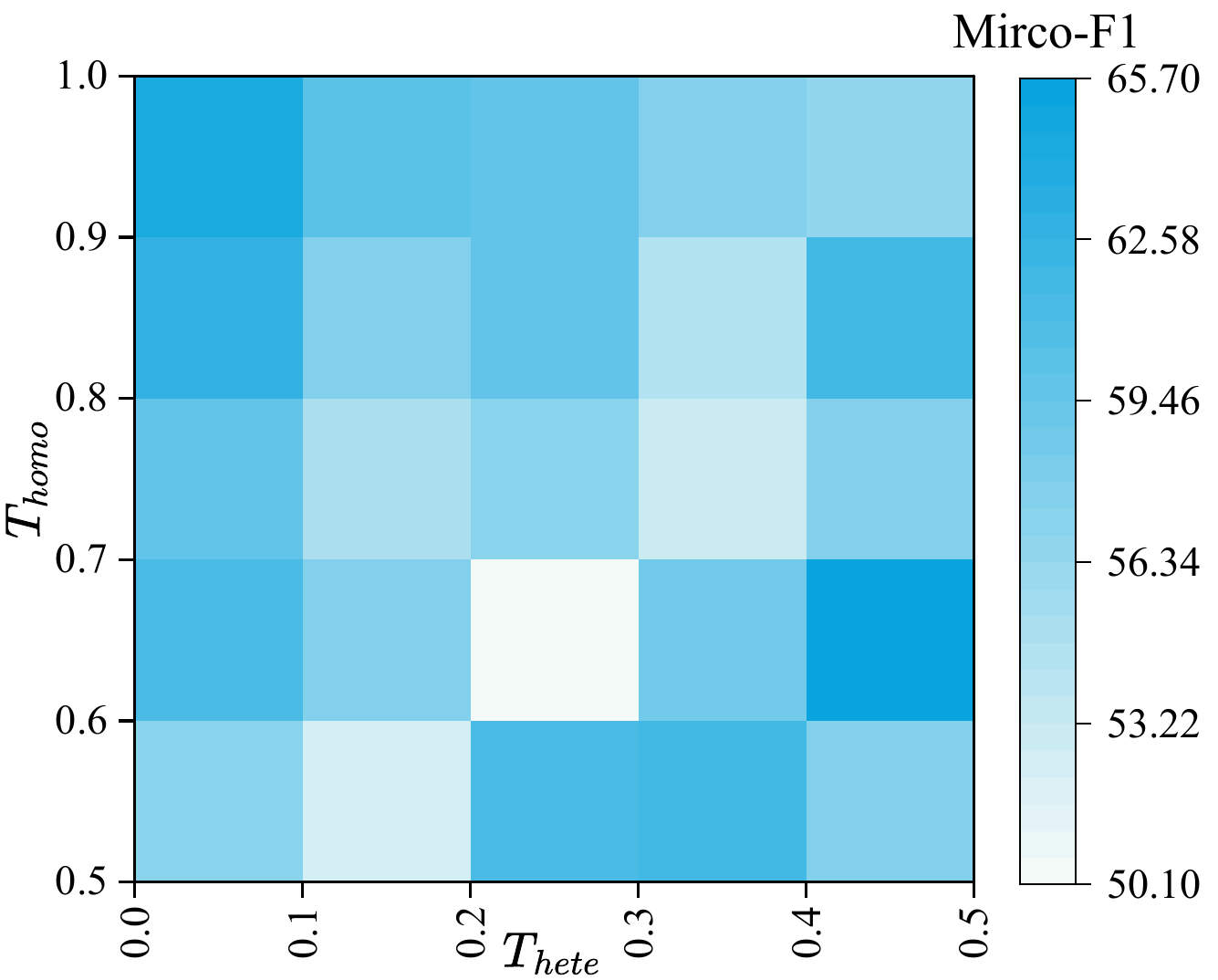}
		\label{wisconsin_thres}}
	\caption{Performance variation when changing hyper-parameters $T_{hete}$ from 0 to 0.5 (step 0.1) and $T_{homo}$ from 0.5 to 1 (step 0.1), with other parameters fixed.}
	\label{ablation_thres}
\end{figure*}  
 
As shown in Figure~\ref{ablation_thres}, the optimal parameter setting on different datasets differs slightly, but is not (0,1), which demonstrates the effectiveness of the heterophily-lessening augmentation.

\begin{figure*}[htbp]
	\centering
	\subfloat[Cora]
	{\includegraphics[width=0.24\linewidth]{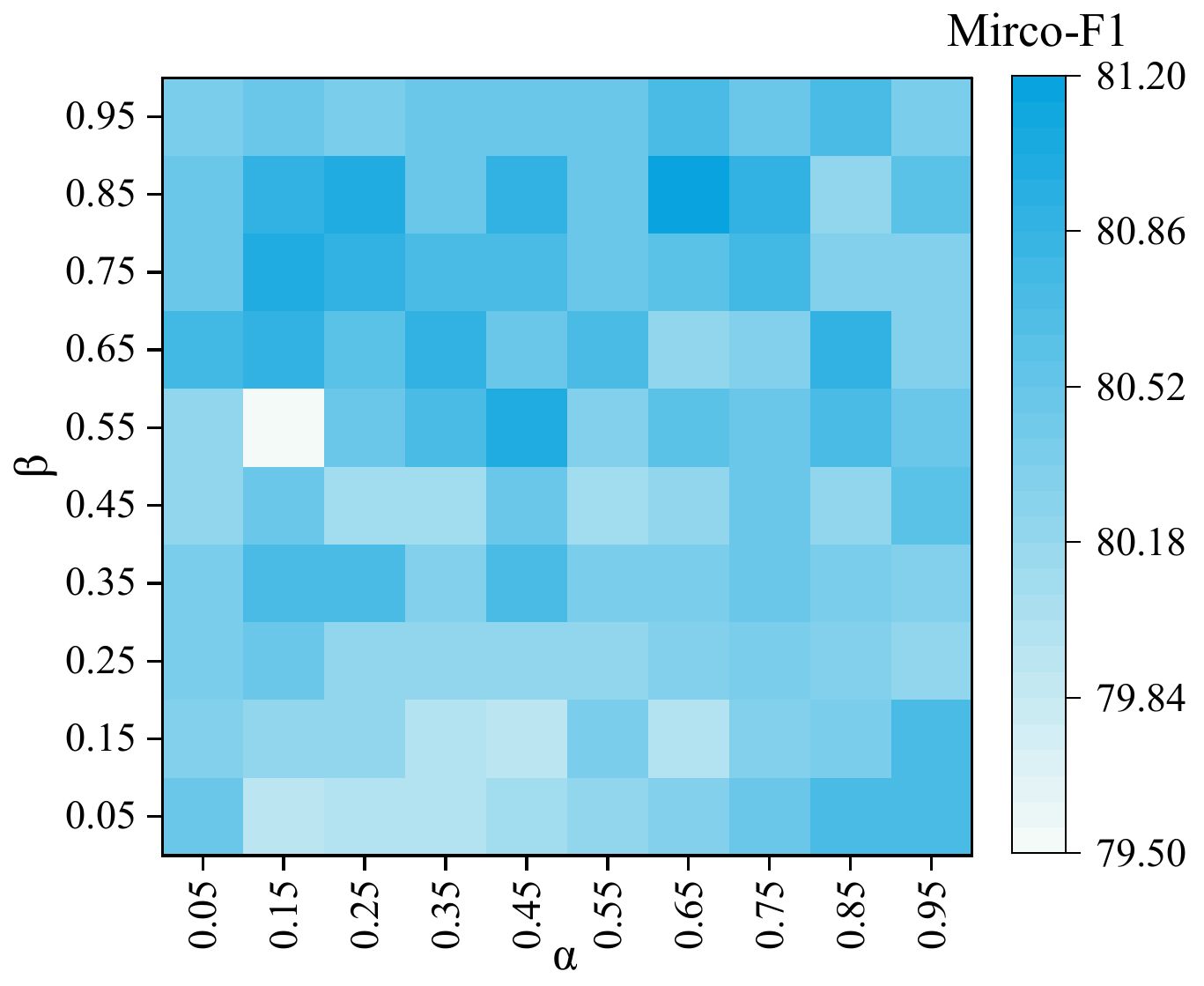}
		\label{cora_ab}}
	\subfloat[CiteSeer]
	{\includegraphics[width=0.24\linewidth]{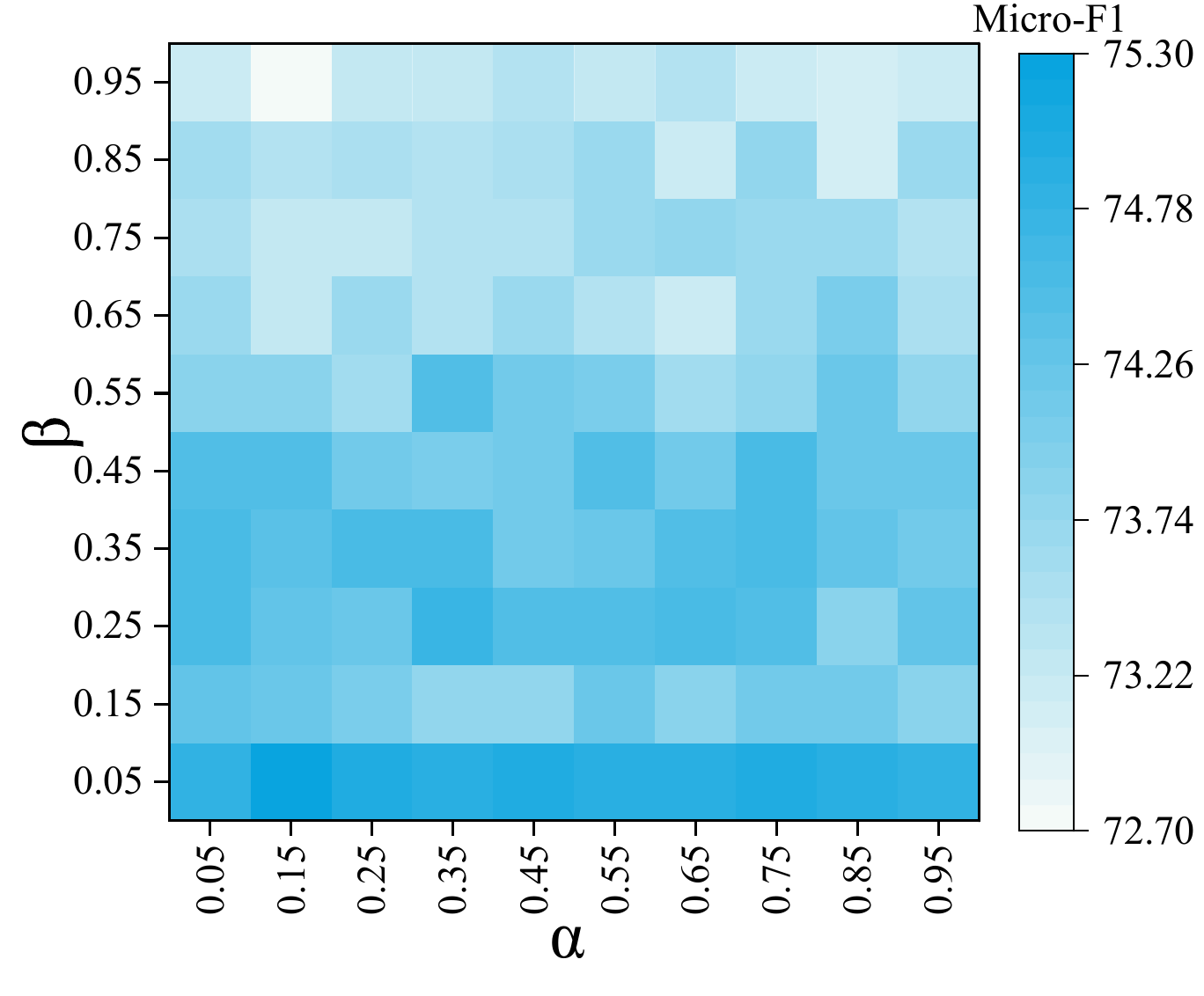}
		\label{citeseer_ab}}
	\subfloat[Chameleon]
	{\includegraphics[width=0.24\linewidth]{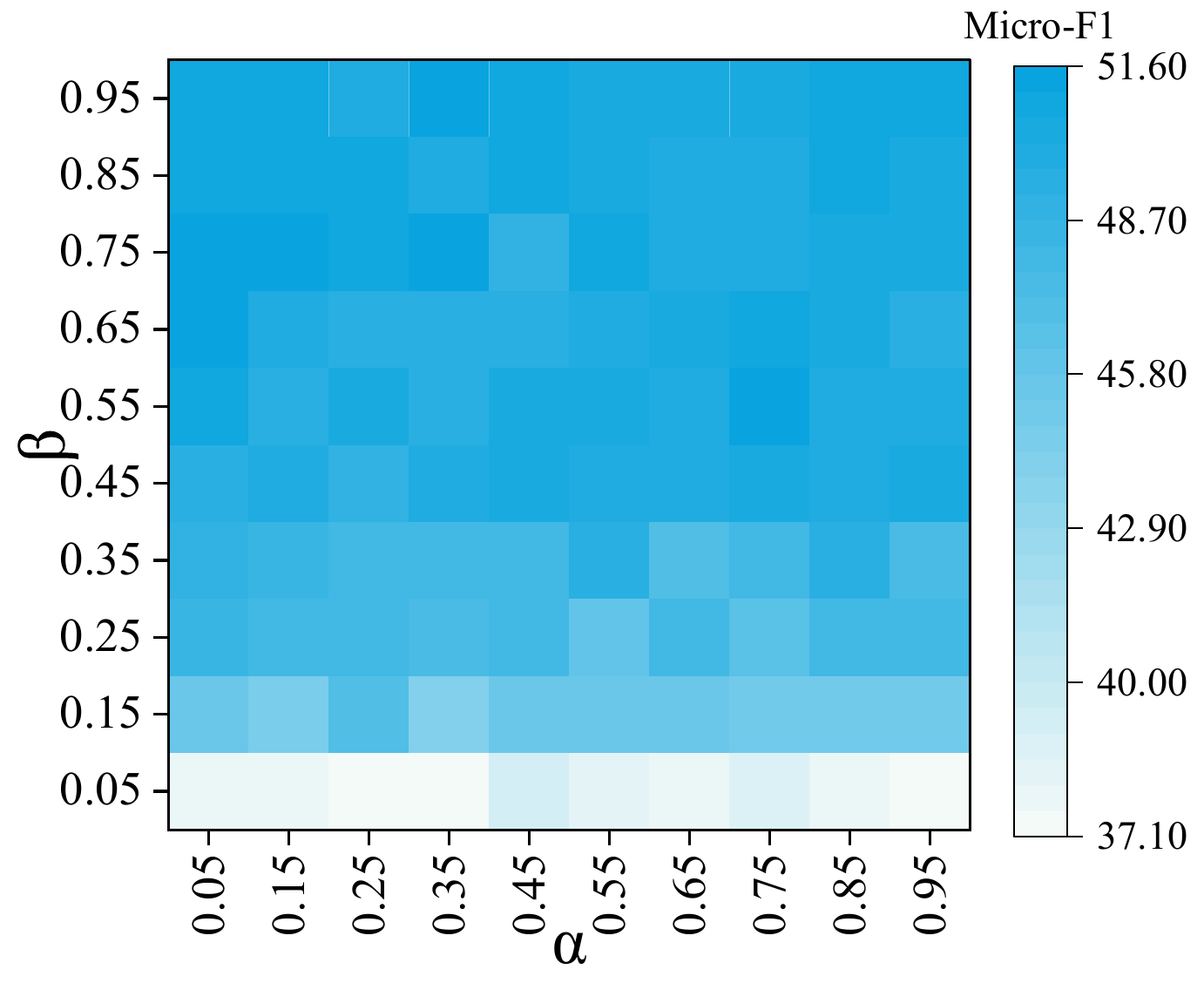}
		\label{chameleon_ab}}
	\subfloat[Squirrel]
	{\includegraphics[width=0.24\linewidth]{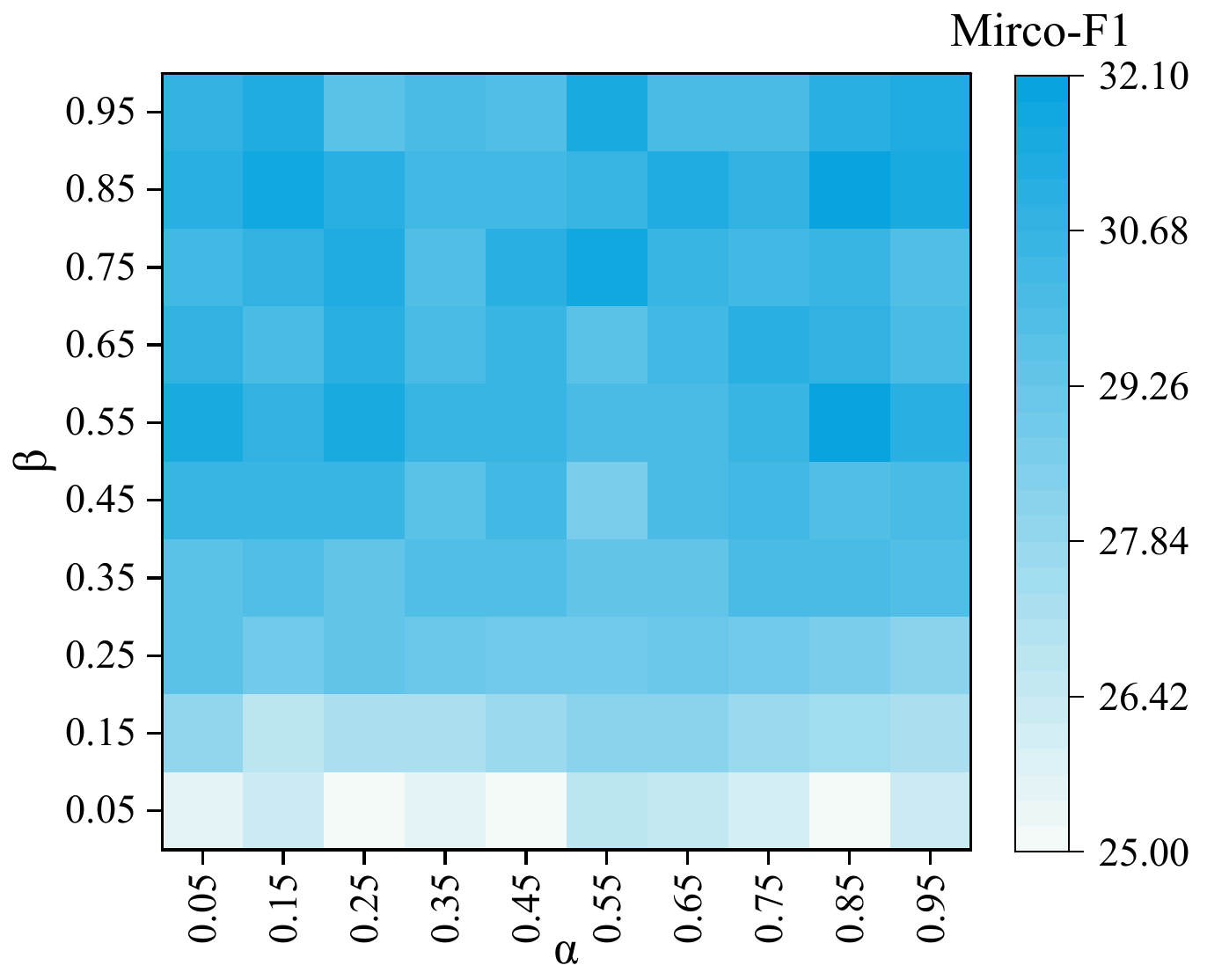}
		\label{squirrel_ab}}
		
	\subfloat[Actor]
	{\includegraphics[width=0.24\linewidth]{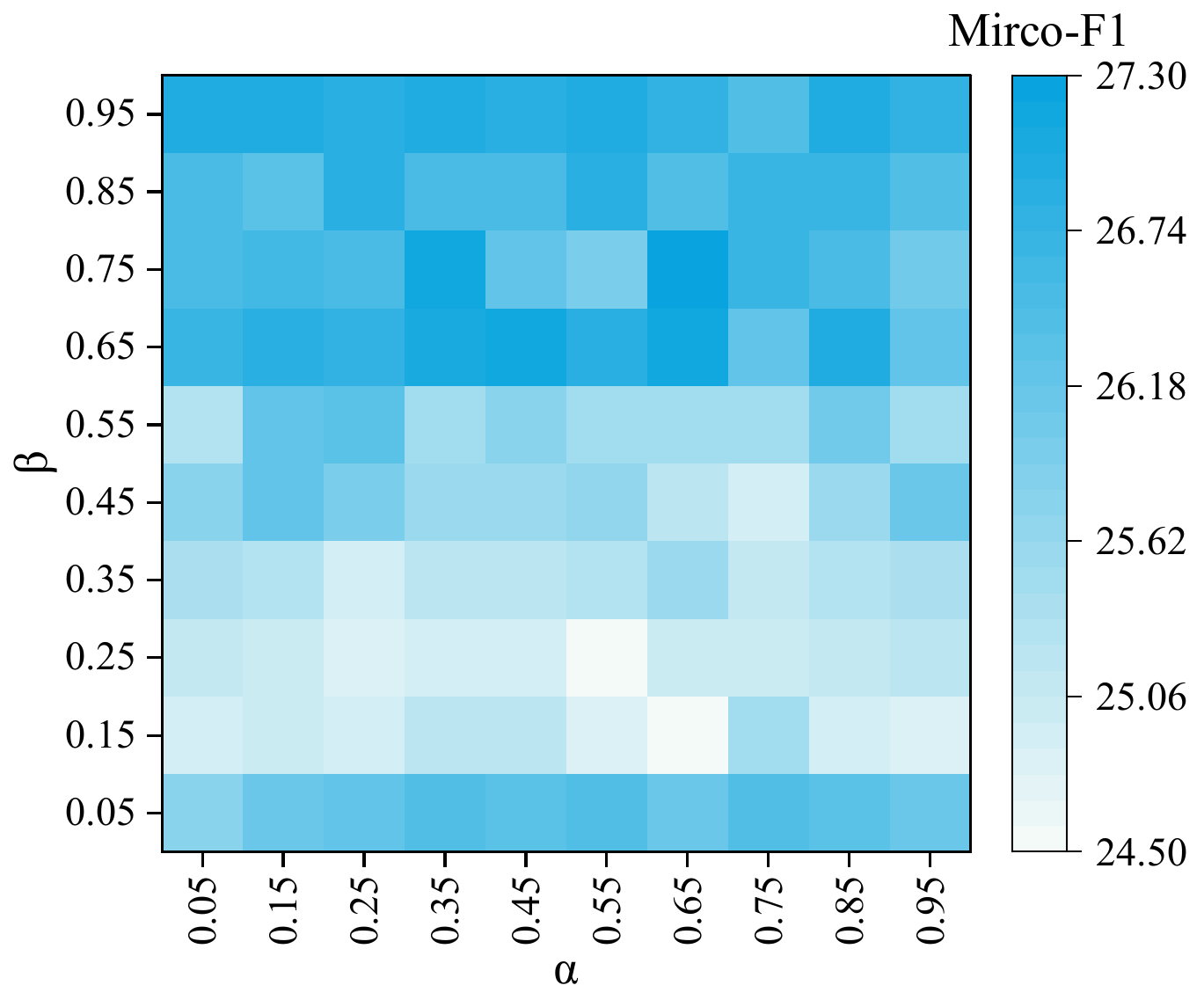}
		\label{actor_ab}}
	\subfloat[Texas]
	{\includegraphics[width=0.24\linewidth]{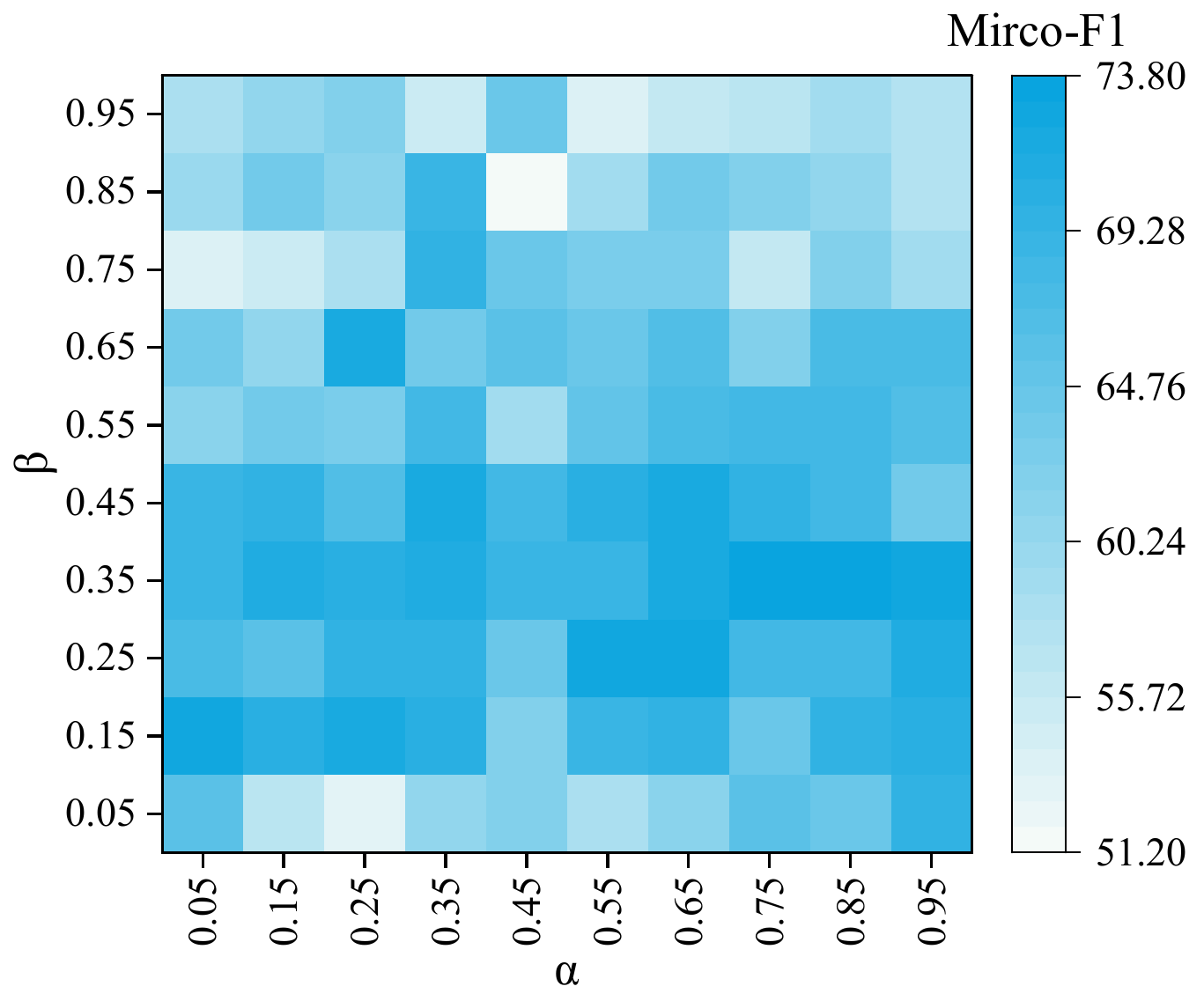}
		\label{texas_ab}}
	\subfloat[Cornell]
	{\includegraphics[width=0.24\linewidth]{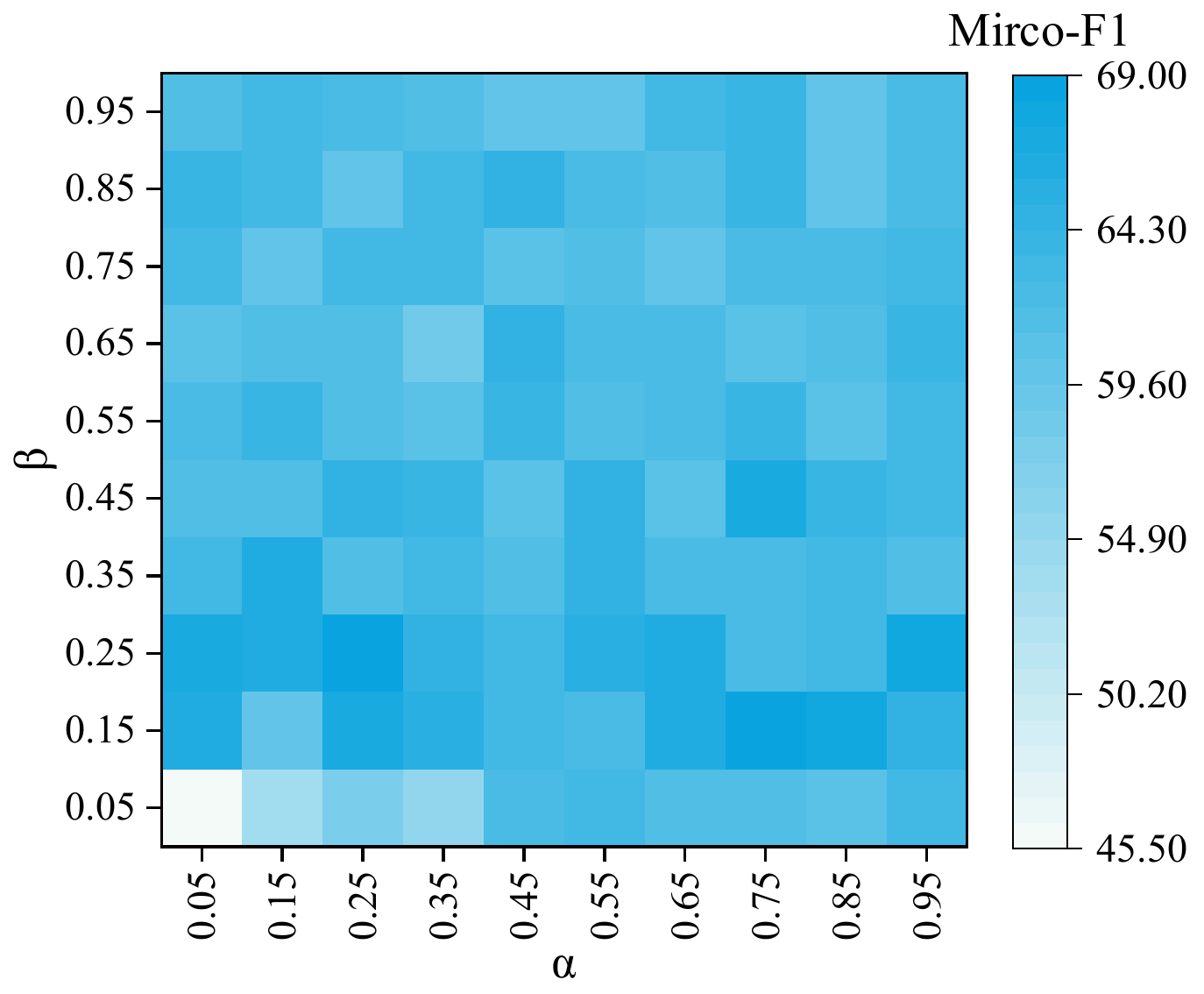}
		\label{cornell_ab}}
	\subfloat[Wisconsin]
	{\includegraphics[width=0.24\linewidth]{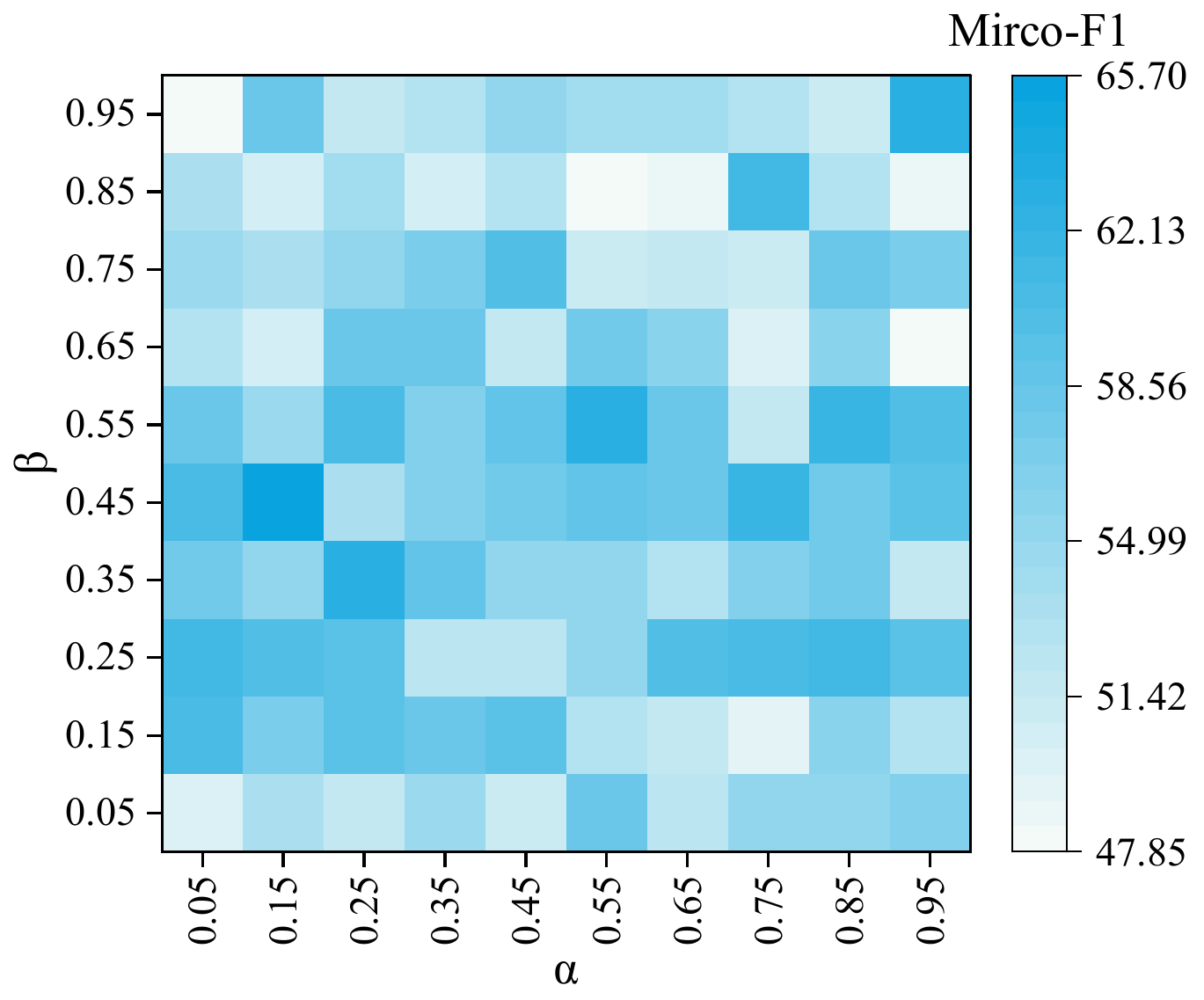}
		\label{wisconsin_ab}}
	\caption{Performance variation when changing hyper-parameters $\alpha$ and $\beta$ from 0.05 to 0.95 (step 0.05), with other parameters fixed.}
	\label{ablation_ab}
\end{figure*}

Figure~\ref{ablation_ab} shows that the model's performance stabilizes with respect to $\alpha$, indicating that both one-hop and two-hop neighbors are important. The optimal value of the parameter $\beta$ exhibits variation across different datasets, indicating that the neighborhood information from homophilic head node plays varying degrees of roles in transferring information to tail nodes across different datasets.

\begin{figure}[htbp]
	\centering
	\subfloat[$k$]
	{\includegraphics[width=0.46\linewidth]{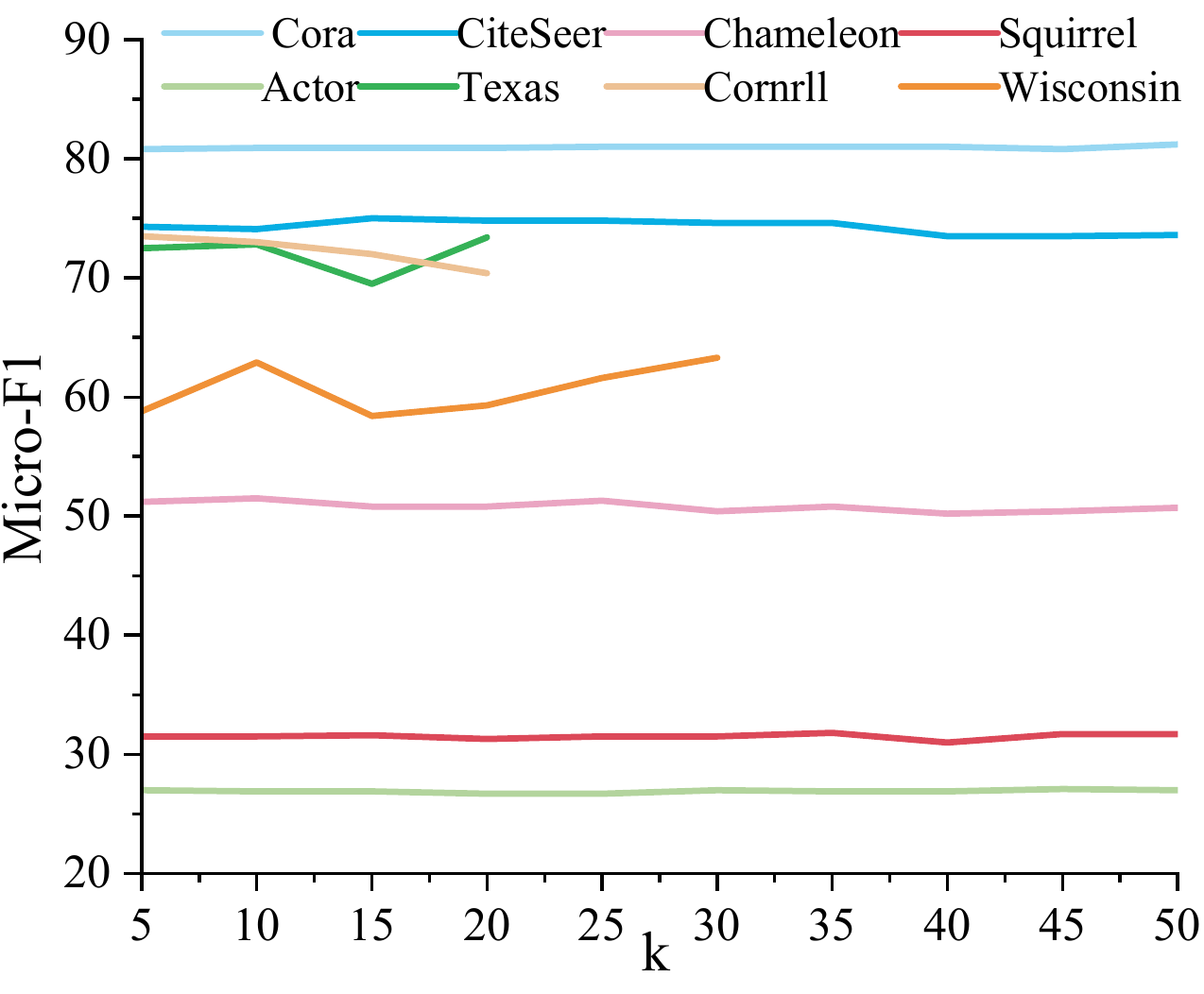}
		\label{topk}}
	\subfloat[$\mu$]
	{\includegraphics[width=0.49\linewidth]{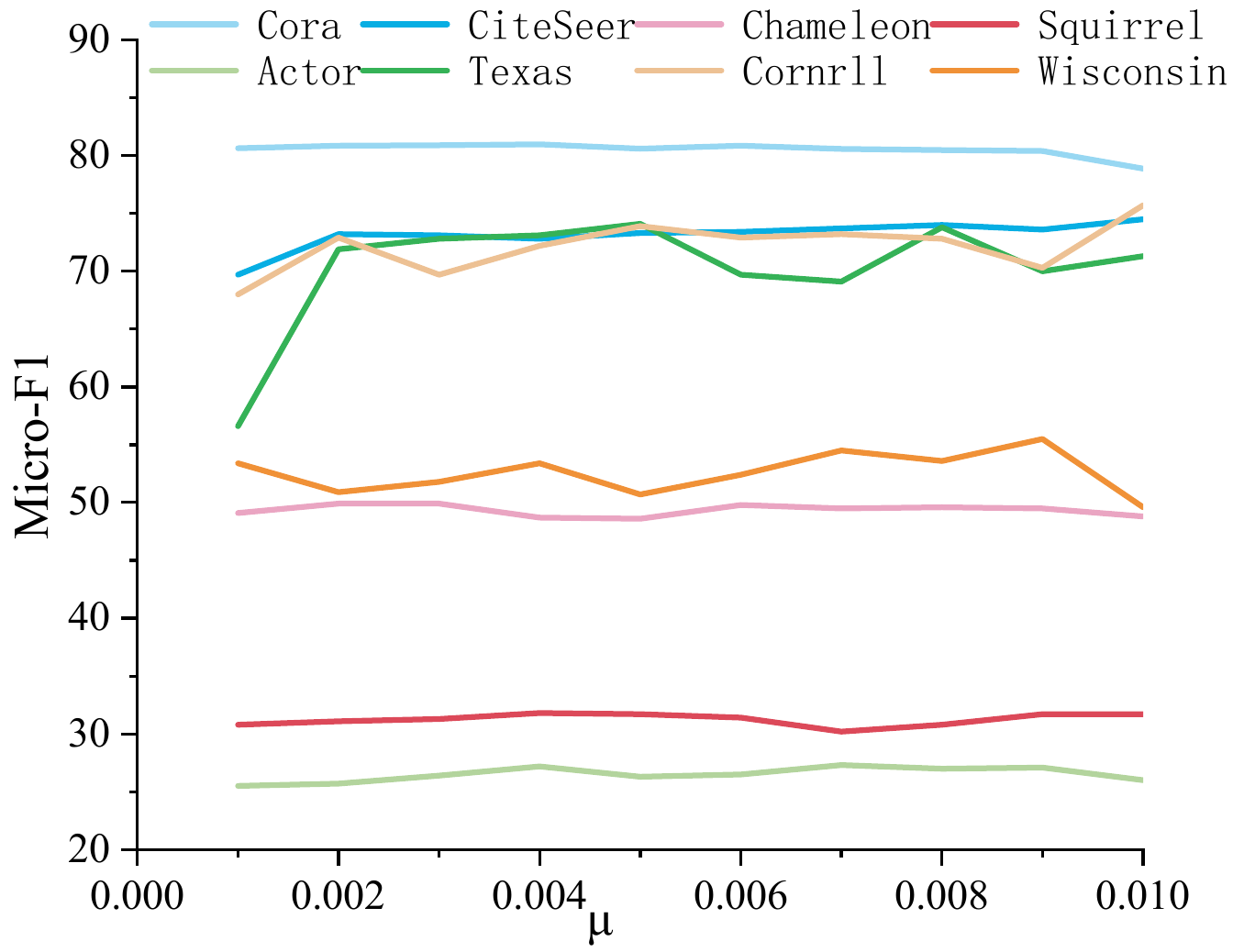}
		\label{mu}}
	\caption{Performance variation when changing hyper-parameters $k$ from 5 to 20 on Cornell and Texas, 30 on Wisconsin, and 50 on other datasets (step 5) and $\mu$ from 0.001 to 0.01 with (step 0.001), with other parameters fixed.}
	\label{ablation_others}
\end{figure} 

The result in Figure ~\ref{ablation_others} shows that for most datasets, $k$ and $\mu$ show little sensitivity. But on small-scale datasets, such as Texas, Wisconsin and Cornell, changes in $k$ cause performance fluctuations due to low node degrees, as excessive expansion of the neighborhood can also introduce noise. Although small-scale datasets are sensitive to any fluctuation, the best value of $\mu$ is not 0, demonstrating the effectiveness of the head node constraint.

\section{Conclusion}
In this paper, we propose HeRB, a novel GNN framework designed to address the structural imbalance issue in heterophilic-aware scenarios. Firstly, through a heterophily-lessening augmentation module, HeRB effectively mitigates heterophilic properties of nodes by increasing intra-class edges and reducing inter-class edges. Subsequently, the homophilic knowledge transfer mechanism enhances the representation of tail nodes by leveraging homophilic information from head nodes, exhibiting more balance during message passing. Experiments on eight datasets show its effectiveness and adaptability across different graphs. Currently, HeRB focuses on transductive scenarios, and future work will be devoted to explore its effectiveness in inductive scenarios.

\section*{Acknowledgments}
This research was supported by the National Natural Science Foundation of China (Grant No. 62476137), the Foundation of State Key Laboratory for Novel Software Technology at Nanjing University (Grant No. KFKT2022B01) and the Natural Science Foundation of Nanjing University of Posts and Telecommunications (Grant No. NY221071).


\bibliographystyle{IEEEtran}
\bibliography{IEEE25.bib}

\end{document}